\documentclass[11pt,english]{article}
\usepackage[latin9]{inputenc}
\usepackage{geometry}
\usepackage{booktabs}
\usepackage{amsmath}
\usepackage{amsthm}
\usepackage{amssymb}
\usepackage{graphicx}

\makeatletter

\providecommand{\tabularnewline}{\\}

\numberwithin{equation}{section}
\numberwithin{figure}{section}
\theoremstyle{plain}
\newtheorem*{prop*}{\protect\propositionname}

\usepackage{amsmath,amsfonts,amssymb}
\usepackage{graphicx}
\usepackage{array}
\usepackage{wrapfig}
\usepackage{tikz}
\usepackage{rotating}
\usepackage{subcaption}

\usepackage{algorithm}
\usepackage{algorithmic}
\usepackage{setspace}
\usepackage{booktabs}
\usepackage{chngcntr}
\usepackage{listings}
\usepackage{color}

\definecolor{mygreen}{rgb}{0,0.6,0}
\definecolor{mygray}{rgb}{0.5,0.5,0.5}
\definecolor{mymauve}{rgb}{0.58,0,0.82}

\counterwithout{figure}{section}

\date{}

\def\E{\mathbb{E}}

\makeatother

\usepackage{babel}
\providecommand{\propositionname}{Proposition}

\begin{document}
\title{Closed-form Expressions for Maximum Mean Discrepancy with Applications
to Wasserstein Auto-Encoders}
\author{Raif M. Rustamov, Data Science and AI Research, AT\&T Labs, Bedminster,
NJ}
\maketitle
\begin{abstract}
The Maximum Mean Discrepancy (MMD) has found numerous applications
in statistics and machine learning, most recently as a penalty in
the Wasserstein Auto-Encoder (WAE). In this paper we compute closed-form
expressions for estimating the Gaussian kernel based MMD between a
given distribution and the standard multivariate normal distribution.
This formula reveals a connection to the Baringhaus-Henze-Epps-Pulley
(BHEP) statistic of the Henze-Zirkler test and provides further insights
about the MMD. We introduce the standardized version of MMD as a penalty
for the WAE training objective, allowing for a better interpretability
of MMD values and more compatibility across different hyperparameter
settings. Next, we propose using a version of batch normalization
at the code layer; this has the benefits of making the kernel width
selection easier, reducing the training effort, and preventing outliers
in the aggregate code distribution. Our experiments on synthetic and
real data show that the analytic formulation improves over the commonly
used stochastic approximation of the MMD, and demonstrate that code
normalization provides significant benefits when training WAEs. 
\end{abstract}

\section{Introduction}

The Maximum Mean Discrepancy (MMD) is a measure of divergence between
distributions \cite{mmd} which has found numerous applications in
statistics and machine learning; see the recent review \cite{MMD_review}
and citations therein. MMD has a well-established theory, based on
which a number of approaches are available for computing the thresholds
for hypothesis testing, allowing to make sense of the raw MMD values;
however, the whole process can be somewhat intricate. Given the increasing
adoption, it is desirable to have closed-form expressions for the
MMD so as to make it more accessible to a general practitioner and
to streamline its use. Additionally, since the raw MMD values are
hard to interpret, it would be important to convert MMD to a more
intuitive scale and provide some easy to remember thresholds for testing
and evaluating model convergence.

We will concentrate on an application of the MMD in the context of
Wasserstein Auto-Encoders, a popular unsupervised learning approach.
MMD quickly entered the neural network arena as a penalty/regularization
term in generative modeling---initially within the moment-matching
generative networks \cite{GMMN1,GMMN2} and later on as a replacement
for the adversarial penalty in Adversarial Auto-Encoders \cite{AdversAE}
leading to the MMD version of Wasserstein Auto-Encoders \cite{InfoVAE,vegan_cookbook,WAE}.
These WAE-MMDs, to which we will refer simply as WAEs, use an objective
that in addition to the reconstruction error includes an MMD term
that pushes the latent representation of data towards some reference
distribution such as multivariate normal. Similarly to Variational
Auto-Encoders \cite{VAE}, WAEs can be used to generate new data samples
by feeding random samples from the reference distribution to the decoder.
By making a fundamental connection to optimal transport distances
in the data space, \cite{vegan_cookbook,WAE} establish theory proving
the correctness of this generative procedure.

Already in the context of WAEs there has been an effort to replace
the MMD with closed-form alternatives. For example, Tabor et al. \cite{Cramer_Wold_AE}
introduce the Cramer-Wold Auto-Encoders inspired by the slicing idea
of \cite{kolouri2018sliced}. While their Cramer-Wold distance has
a closed-form expression, it depends on special functions unless one
uses an approximation. In addition, similarly to the situation with
the MMD, the raw values of the Cramer-Wold distance are not directly
interpretable.

In this paper, we carry out the analytical computation of the MMD
in a special case where the reference distribution is the standard
multivariate normal and the MMD kernel is a Gaussian RBF. We are also
able to compute the variance of the MMD in closed-form, which allows
us to introduce the standardized version of the MMD. Our MMD formula
reveals a relationship to the Baringhaus-Henze-Epps-Pulley (BHEP)
statistic \cite{EP,BH} used in the Henze-Zirkler test of multivariate
normality \cite{HZ}, which allows making a connection to the Cramer-Wold
distance.

Focusing on WAEs as an application, we discuss the use of the closed-form
standardized MMD as a penalty in the WAE training objective. Estimating
the MMD the usual way requires sampling both from the latent code
and the target reference distributions. The latter sampling incurs
additional stochasticity which has an immediate effect on the gradients
for training; using the analytic formula for the MMD integrates out
this extra stochasiticty. The standardization of the MMD induces better
compatibility across different hyperparameter settings, which can
be advantageous for model selection. In addition, it is more amenable
to direct interpretation, which is demonstrated by easy to remember
rules of thumb suitable for model evaluation. As another contribution,
we propose using code normalization--- a version of the batch normalization
\cite{batchnorm} applied at the code layer---when training WAEs.
This has the benefits of making the selection of width for the MMD
kernel easier, reducing the training effort, and preventing outliers
in the aggregate latent distribution. 

The paper is organized as follows. Section \ref{sec:Closed-Form-Expressions}
provides closed-form expressions for the MMD and discusses the connections
to BHEP statistic. In Section \ref{sec:WAE-Training}, we provide
a number of suggestions for training WAEs and monitoring the training
progress. Section \ref{sec:Experiments} provides an empirical evaluation
on synthetic and real data. The derivations of the formulas, extensions,
and relevant code are provided in the Appendix.

\section{\label{sec:Closed-Form-Expressions}Closed-Form Expressions for MMD}

The maximum mean discrepancy is a divergence measure between two distributions
$P$ and $Q$. In the context of WAEs, applying the encoder net to
the distribution of the input data (e.g. images) yields the aggregate
distribution $Q$ of the latent variables. One of the goals of WAE
training is to make $Q$ (which depends on the neural net parameters)
as close as possible to some fixed target distribution $P$. This
is achieved by incorporating MMD between $P$ and $Q$ as a regularizer
into the WAE objective.

The computation of the MMD requires specifying a positive-definite
kernel; in this paper we always assume it to be the Gaussian RBF kernel
of width $\gamma$, namely, $k(x,y)=e^{-\Vert x-y\Vert^{2}/(2\gamma^{2})}$.
Here, $x,y\in\mathbb{R}^{d}$, where $d$ is the dimension of the
code/latent space, and we use $\Vert\cdot\Vert$ to denote the $\ell_{2}$
norm. The population MMD can be most straight-forwardly computed via
the formula \cite{mmd}:

\begin{equation}
\mathrm{MMD}^{2}(P,Q)=\E_{x,x'\sim P}[k(x,x')]-2\E_{x\sim P,y\sim Q}[k(x,y)]+\E_{y,y'\sim Q}[k(y,y')].\label{eq:mmd}
\end{equation}
In this paper, the target reference distribution $P$ is always assumed
to be the standard multivariate normal distribution $\mathcal{N}_{d}$
with the density $p(x)=(2\pi)^{-d/2}e^{-\Vert x\Vert^{2}/2}$, $x\in\mathbb{R}^{d}$;
to simplify the formulas we will use the short-hand notation $\mathcal{N}_{d}$.
In practical situations, we only have access to $Q$ through a sample.
For example, during each step of the WAE training, the encoder neural
net will compute the codes $z_{i},i=1,...,n$ corresponding to the
input data in the batch (we use ``batch'' to mean ``mini-batch'')
and the current values of neural network parameters. Given this sample
from $Q$, our goal is to derive a closed-form estimate of $\mathrm{MMD^{2}}(P,Q)$.

\paragraph*{Unbiased Estimator}

We start with the expression Eq. (\ref{eq:mmd}) and using the sample
$Q_{n}=\{z_{i}\}_{i=1}^{n}$ of size $n$, we replace the last two
terms by the sample average and the U-statistic respectively to obtain
the unbiased estimator \cite{mmd}:

\begin{equation}
\mathrm{MMD}_{u}^{2}(\mathcal{N}_{d},Q_{n})=\E_{x,x'\sim\mathcal{N}_{d}}[k(x,x')]-\frac{2}{n}\sum_{i=1}^{n}\E_{x\sim\mathcal{N}_{d}}[k(x,z_{i})]+\frac{1}{n(n-1)}\sum_{i=1}^{n}\sum_{j\neq i}^{n}k(z_{i},z_{j}).\label{eq:mmdu}
\end{equation}
Our main result is the following proposition whose proof can be found
in Appendix \ref{sec:Derivations-and-Proofs}:
\begin{prop*}
The expectations in the expression above can be computed analytically
to yield the formula:

\[
\mathrm{MMD}_{u}^{2}(\mathcal{N}_{d},Q_{n})=\left(\frac{\gamma^{2}}{2+\gamma^{2}}\right)^{d/2}-\frac{2}{n}\left(\frac{\gamma^{2}}{1+\gamma^{2}}\right)^{d/2}\sum_{i=1}^{n}e^{-\frac{\Vert z_{i}\Vert^{2}}{2(1+\gamma^{2})}}+\frac{1}{n(n-1)}\sum_{i=1}^{n}\sum_{j\neq i}^{n}e^{-\frac{\Vert z_{i}-z_{j}\Vert^{2}}{2\gamma^{2}}}.
\]
 In addition, we can compute the variance under the null $Q=P=\mathcal{N}_{d}$,

\begin{align}
\mathrm{Var}(\gamma,d,n)\triangleq & \mathrm{Var}_{Q_{n}\sim\mathcal{N}_{d}}\left[\mathrm{MMD_{u}^{2}}(\mathcal{N}_{d},Q_{n})\right]=\nonumber \\
= & \frac{2}{n(n-1)}\left[\left(\frac{\gamma^{2}}{2+\gamma^{2}}\right)^{d}+\left(\frac{\gamma^{2}}{4+\gamma^{2}}\right)^{d/2}-2\left(\frac{\gamma^{4}}{(1+\gamma^{2})(3+\gamma^{2})}\right)^{d/2}\right].\label{eq:Variance}
\end{align}
\end{prop*}
Having a closed-form formula for the MMD is advantageous for optimization
of WAEs. Computing this penalty the usual way \cite{mmd,WAE,InfoVAE}
relies on taking a sample from both $P=\mathcal{N}_{d}$ and $Q$.
As a result, this incurs additional stochasticity due to the sampling
from $P$. Our formula essentially integrates out this stochasticity,
and results in an estimator with a smaller variance. This allows better
discrimination between distributions (see Section \ref{sec:Experiments}),
and, as a result, potentially provides higher quality gradients for
training. In the Appendix we also provide closed-form formulas for
the random encoder variant of the WAE.

On a conceptual level, this formula for $\mathrm{MMD}_{u}^{2}$ reveals
two forces at play when optimizing $Q_{n}$ to have small divergence
from the standard multi-variate normal distribution: one force is
pulling the sample points towards the origin, and the other is pushing
them apart from each other. Another observation is that one can compute
the optimal translation transform for a given sample, and surprisingly
it is not the one that places the center of mass at the origin. In
fact, during this shift optimization the third term stays constant,
and the second term can be interpreted (up-to a constant factor) as
a kernel density estimate with the kernel width of $1+\gamma^{2}$.
The optimal shift is the one that places the mode of this density
estimate at the origin.

\paragraph*{Biased Estimator and BHEP Statistic}

The biased estimator from \cite{mmd} can be computed in closed form
in a similar manner. The only difference is the use of the V-statistic
for the third term in Eq. (\ref{eq:mmd}); the final expression is
as follows:

\[
\mathrm{MMD}_{b}^{2}(\mathcal{N}_{d},Q_{n})=\left(\frac{\gamma^{2}}{2+\gamma^{2}}\right)^{d/2}-\frac{2}{n}\left(\frac{\gamma^{2}}{1+\gamma^{2}}\right)^{d/2}\sum_{i=1}^{n}e^{-\frac{\Vert z_{i}\Vert^{2}}{2(1+\gamma^{2})}}+\frac{1}{n^{2}}\sum_{i=1}^{n}\sum_{j=1}^{n}e^{-\frac{\Vert z_{i}-z_{j}\Vert^{2}}{2\gamma^{2}}}.
\]
Interestingly, this expression is equivalent to a statistic proposed
for testing multivariate normality, its history going back to as early
as 1983. The Baringhaus-Henze-Epps-Pulley (BHEP) statistic is named
after \cite{BH} and \cite{EP} as coined by \cite{Csorgo}. This
statistic is used in the Henze-Zirkler test of multivariate normality
\cite{HZ}. 

The BHEP statistic is a measure of divergence between two distributions
$P$ and $Q$ that captures how different their characteristic functions
are. It is defined as the weighted $L^{2}$-distance:
\[
W(P,Q)=\int_{\mathbb{R}^{d}}\vert\Psi^{P}(t)-\Psi^{Q}(t)\vert^{2}\varphi(t)dt
\]
where $\Psi^{P}(t)$ and $\Psi^{Q}(t)$ are the characteristic functions
of the distributions $P$ and $Q$, and $\varphi(t)$ is a weight
function. When $P$ is the standard multivariate normal distribution
$\mathcal{N}_{d}$, we have $\Psi^{P}(t)=\exp(-\Vert t\Vert^{2}/2)$.
Selecting the weight function to be $\varphi_{\beta}(t)=(2\pi\beta^{2})^{-d/2}e^{-\Vert x\Vert^{2}/(2\beta^{2})}$
and and replacing $Q$ by a sample $Q_{n}$, the BHEP statistic takes
the following form:

\[
W_{n,\beta}=\int_{\mathbb{R}^{d}}\vert\exp(-\Vert t\Vert^{2}/2)-\Psi^{Q_{n}}(t)\vert^{2}\varphi_{\beta}(t)dt.
\]
Here $\Psi^{Q_{n}}(t)$ is the empirical characteristic function of
$Q$,
\[
\Psi^{Q_{n}}(t)=\frac{1}{n}\sum_{i=1}^{n}\exp(\sqrt{-1}\:t\cdot z_{i}),\;t\in\mathbb{R}^{d}.
\]
A closed-form formula for $W_{n,\beta}$ can be obtained (see e.g.
\cite{HZ,HenzeWagner1997}) and \emph{it coincides with the expression
for} $\mathrm{MMD}_{b}^{2}$ \emph{when one sets the Gaussian RBF
kernel width} $\gamma=1/\beta$. 

This connection has a number of useful consequences. First, inspecting
the relationship between the MMD and the characteristic function formulation
of the BHEP statistic, we see that this formulation more transparently
expresses the fact that MMD is performing moment matching; curiously,
the formula for $W_{n,\beta}$ provides a connection to the Random
Fourier Features \cite{Rahimi:2007:RFL:2981562.2981710} and their
use for the MMD computation \cite{Fastmmd}. Second, \cite{HZ} show
that BHEP statistic can be equivalently obtained as the $L^{2}$-distance
between kernel density estimates; in our context, this is a concrete
example of the connection described in \cite[Section 3.3.1]{mmd}.
Third, based on this equivalence, \cite{HZ} suggests using a specific
value of $\beta$ from optimal density estimation theory. The corresponding
$\gamma$ is $\gamma_{d,n}=1/\beta_{d,n}=\sqrt{2}\left((2d+1)n/4\right)^{-1/(d+4)}$
to which we will refer as HZ $\gamma$ in our experimental section.
Fourth, the one-dimensional distance used in the definition of the
Cramer-Wold distance \cite{Cramer_Wold_AE} is based on exactly the
same $L^{2}$-distance between kernel density estimates. As a result,
we see that the Cramer-Wold distance is the integral of $\mathrm{MMD}_{b}^{2}$
over all one-dimensional projections of $Q$. Of course, by similarly
integrating $\mathrm{MMD}_{u}^{2}$ instead, one could introduce a
new version of the Cramer-Wold distance that is zero centered under
the null. 

\emph{Remark:} We leave out the computation of the null mean and variance
of the $\mathrm{MMD}_{b}^{2}$; this can be carried out similarly
to $\mathrm{MMD}_{u}^{2}$. Note that the mean and variance of $W_{n,\beta}$
are computed in closed-form in \cite{HZ}. However, these expressions
are based on a composite null hypothesis and have corrections for
nuisance parameter estimation. 

\section{\label{sec:WAE-Training}Suggestions for WAE Training }

\subsection{\label{sec:Scaled-Closed-Form-MMD}Standardized MMD Penalty}

\begin{figure}
\includegraphics[width=1\textwidth]{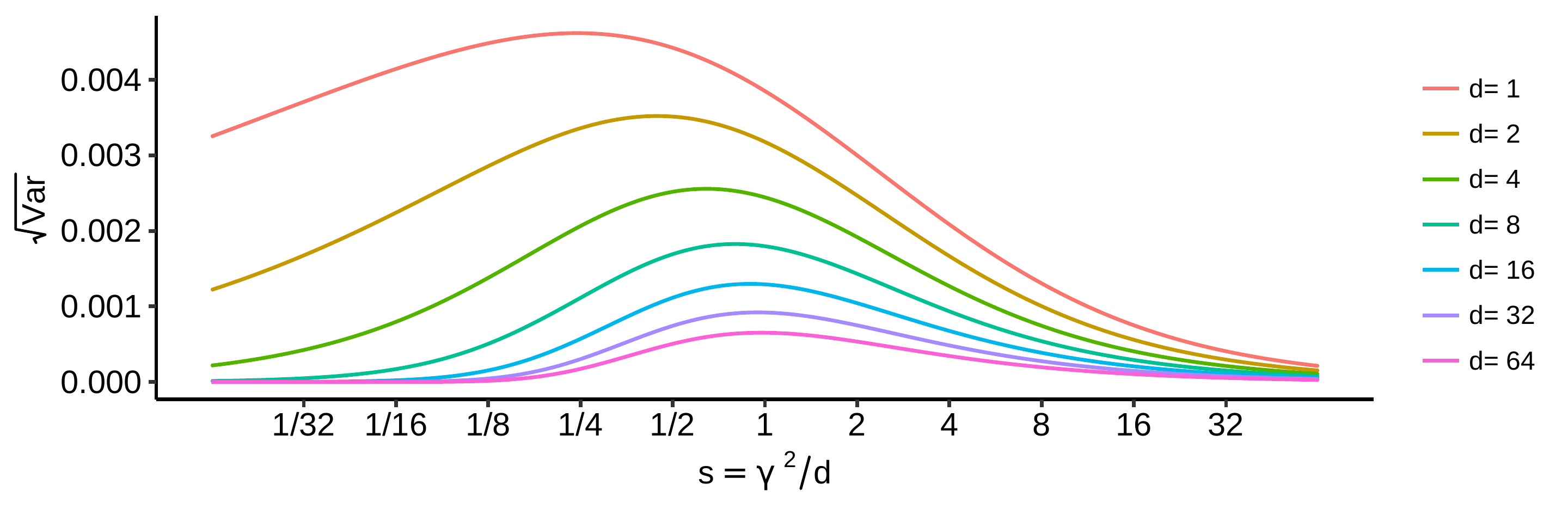}

\caption{\label{fig:Variance-curve}Variance as a function of kernel width
$\gamma$ and latent dimensionality $d$. Batch size is fixed to $n=100$.}
\end{figure}
In the original formulation of the WAE, the MMD penalty enters the
objective as the term $\lambda\cdot\mathrm{MMD}^{2}(\mathcal{N}_{d},Q)$,
where $\lambda$ is the regularization strength. Obviously, the closed-form
formulas for the MMD presented in the previous section can be used
instead. In addition, we suggest standardizing the MMD. Since the
mean of $\mathrm{MMD}_{u}^{2}$ under the null is zero, and variance
under the null is $\mathrm{Var}(\gamma,d,n)$ as given by Eq. (\ref{eq:Variance}),
we define:

\[
\mathrm{SMMD}^{2}(Q_{n})\triangleq\frac{\mathrm{MMD}_{u}^{2}(\mathcal{N}_{d},Q_{n})}{\sqrt{\mathrm{Var}(\gamma,d,n)}}.
\]
Figure \ref{fig:Variance-curve} depicts the behavior of the scaling
term $\sqrt{\mathrm{Var}(\gamma,d,n)}$ for different values of the
kernel width $\gamma$ and the latent dimension $d$, for a fixed
batch size $n=100$. We can clearly see that the scaling varies widely
not only across dimensions, but also across kernel width choices at
a fixed dimension. 

While at the theoretical level the suggested scaling can be equivalently
seen as a re-definition of the regularization coefficient, yet it
has a number of benefits in practice. First, the use of the SMMD is
potentially beneficial for model selection.The choice of the best
hyperparameters is usually carried out via cross-validation which
among others things includes trying out different values of the penalty
coefficient $\lambda$, kernel width $\gamma$, and latent dimension
$d$. Without the proposed scaling of the MMD term, the values of
$\lambda$ are not universal across the choices of $\gamma$ and $d$.
For example, if a small list of $\lambda$'s is used when cross-validating,
then disparate regions of the optimization space would be considered
across the choices of $\gamma$ and $d$, perhaps resulting in a suboptimal
model being chosen. Second, our scaled formulation can also be beneficial
for the commonly used trick of combining kernels of different widths---using
a penalty of the form $\lambda\cdot[\mathrm{MMD}^{2}(\mathcal{N}_{d},Q,\gamma=\gamma_{1})+...+\mathrm{MMD}^{2}(\mathcal{N}_{d},Q,\gamma=\gamma_{k})]$---in
order to boost the performance of the MMD and to avoid search over
the kernel width. However, when such a combination is performed without
the proposed standardization, then MMDs coming from different kernel
width choices can be of different orders of magnitude. As a result,
one may end up with a single kernel width dominating. In fact, the
common choice of including kernels having $s=\gamma^{2}/d\approx1$
together with the ones that have $s=\gamma^{2}/d\ll1$ or $\gg1$
would lead to this issue as can be seen from Figure \ref{fig:Variance-curve}.
One can see that this observation is also relevant in cases where
$\gamma$ is set adaptively per batch, this time leading to various
amounts of penalty being applied to each batch. Third, the most important
advantage of using the SMMD is that it is more interpretable and so
amenable to quick inspection when one wants to have a sense of how
far the current distribution is from the target normal multivariate
distribution; this is the focus of the following discussion.

\paragraph*{Monitoring WAE Training Progress}

It is a standard practice when training a neural network to monitor
the progress by inspecting the total loss and its components both
for training and validation data. These metrics of interest are computed
on a batch level and some type of running averages over the batches
are reported. We consider two types of averaging, simple averaging
and exponential moving averaging of the SMMD values. For both of these
cases we explain how to asses the convergence of code distribution
to the standard multivariate normal. As a complementary approach,
Appendix \ref{sec:Hypothesis-Testing} provides thresholds that can
be used when monitoring convergence on a single batch level without
averaging.

In the case of the simple averaging, the asymptotic distribution under
the null is easy to compute. Assume that the validation set contains
$m$ batches of size $n$, and the corresponding batches are $Q_{n}^{b}=\{z_{i}^{b}\}_{i=1}^{n},b=1,2,...,m$.
The average SMMD value is computed as $B_{m}=\frac{1}{m}\sum_{b=1}^{m}\mathrm{SMMD}^{2}(Q_{n})$,
and is the average of independent and identically distributed terms.
Under the null, each summand has zero mean and unit variance due to
the standardization. Assuming that $m$ is big enough, we can apply
the Central Limit Theorem \cite{CLT_Ref}, giving that the null distribution
of $B_{m}$ is asymptotically normal with mean $0$ and variance $1/m$.
Thus, as a rule of thumb, values of $B_{m}$ that do not fall into
the three-sigma interval $[-3/\sqrt{m},3/\sqrt{m}]$ should be considered
as an indication that the aggregate code distribution has not converged
to the target standard multivariate normal distribution. The raw MMD
version of this test together with theoretical results can be found
in \cite{B_test}, but it is our standardization that makes the test
easily applicable by practitioners. \cite{B_test} called this the
B-test, so we will refer to $B_{m}$ as the B-Statistic.

Another popular way of keeping track of progress metrics is exponential
moving averaging. The Lyapunov/Lindeberg version of the Central Limit
Theorem \cite[Chapter 27]{CLT_Ref} can be applied to obtain the corresponding
interval. Suppose that the exponential moving average with the momentum
of $\alpha$ is used to keep track of a per-batch quantity $S_{b}$.
Thus, $E_{b}=\alpha E_{b-1}+(1-\alpha)S_{b}$ is used for $b=1,...,m$.
Note that, $E_{m}$ can be written as $E_{m}=\alpha^{m}E_{0}+(1-\alpha)[\alpha^{m-1}S_{1}+\alpha^{m-2}S_{2}+\cdots\alpha S_{m-1}+S_{m}],$
here $S_{0}$ is some initial value, usually $0$, which we will use.
Assuming that $S_{b}$ are standardized to have zero mean and unit
variance, the application of the CLT to random variables $(1-\alpha)\alpha^{m-k}S_{k}$
gives that $E_{m}$ is normally distributed: $E_{m}\sim\mathcal{N}(0,(1-\alpha^{2m+2})\frac{1-\alpha}{1+\alpha}).$
By dropping $(1-\alpha^{2m+2})$, we can use $(1-\alpha)/(1+\alpha)$
as an \emph{upper bound} for the variance. This gives the three-sigma
interval for the E-Statistic liberally as $[-3\sqrt{(1-\alpha)/(1+\alpha)},3\sqrt{(1-\alpha)/(1+\alpha)}]$.
For a common value of $\alpha=0.99$ we get the interval as $[-0.212,0.212]$.

\subsection{\label{subsec:Code-Normalization}Code Normalization}

In this subsection we propose to apply a variant of batch normalization
\cite{batchnorm} on top of the code layer before MMD is computed:
for each batch, we center and scale the codes so that their distribution
has zero mean and unit variance in each dimension; we will refer to
this as ``code normalization''. Importantly, no scaling or shifting
is applied after normalizing (i.e. $\gamma=1$ and $\beta=0$ in the
notation of \cite{batchnorm}) as the decoder network expects a normally
distributed input. Below under separate headings we discuss the benefits
of code normalization for the WAE training; we will use the term ``MMD
penalty'' to refer to any kind of penalty based on MMD, including
SMMD. 

\paragraph*{Easier Kernel Width Selection}

One advantage of code normalization is that a single setting of the
width for the Gaussian RBF kernel, $\gamma$, can be used when computing
the MMD penalty. Without code normalization, a fixed choice of $\gamma$
leads to issues. For example, when $\gamma$ is small, and the codes
are far away from the origin and from each other, the MMD penalty
term has small gradients, which makes learning difficult or even impossible.
Indeed, the exponentials become vanishingly small, and since they
enter the gradient multiplicatively this makes the gradients small
as well. The same issue arises when choosing a large value of $\gamma$
when the codes are not far away from the origin. Thus, one has to
use an adaptive choice of $\gamma$ in order to deal with this problem,
see e.g. \cite{WAE}. On the other hand, in the long run, code normalization
makes sure that the codes have commensurate distances with $\gamma$
throughout the training process, alleviating the need for an adaptive
$\gamma$. This makes possible to decouple the choice of $\gamma$
from the neural network training and to provide practical recommendations
as we do in Section \ref{sec:Experiments}.

\paragraph*{Reduced Training Effort}

\begin{figure}
\begin{centering}
\includegraphics[width=1\textwidth]{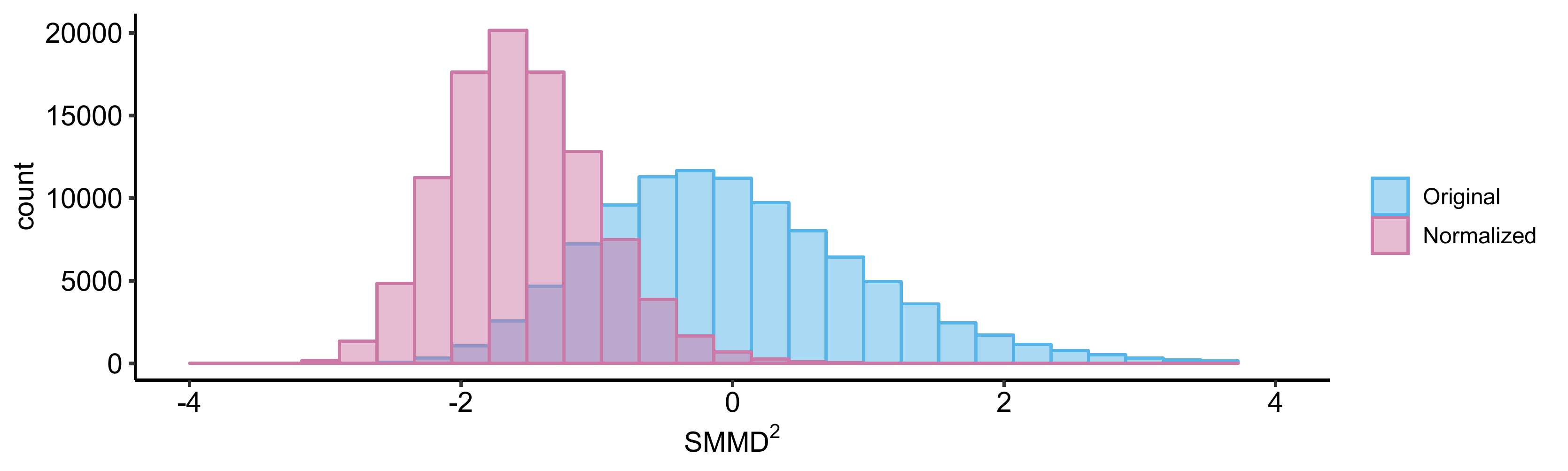}
\par\end{centering}
\caption{\label{fig:codenorm-effect}Code normalization shifts the distribution
of $\mathrm{SMMD}^{2}$ to lower values. Here, $n=100$, $d=8$, and
kernel scale $s=\gamma^{2}/d=1/4$.}
\end{figure}

Code normalization shifts and scales codes to be in the ``right''
part of the space, namely where the target standard multivariate normal
distribution lives, and we speculate that this reduces the training
effort. The intuition comes from inspecting the relationship between
the MMD and the characteristic function formulation of the BHEP statistic.
This formulation expresses the fact that at some level MMD is performing
moment matching, and so by rendering the first two moments (marginal)
equal to those of the standard multivariate normal distribution, code
normalization focuses the training effort on matching the higher moments. 

To illustrate this point, Figure \ref{fig:codenorm-effect} shows
the distribution of SMMD values for samples of size $n=100$ taken
from $Q=\mathcal{N}_{d}$. The value of SMMD is computed for the original
sample and then for the sample to which code normalization was applied.
Note that on average the normalized codes have smaller MMD values
compared to the original ones. In a sense, normalized samples are
more ``ideal'' from the point of the view of the MMD. This means
that even if the neural network has converged to the target normal
distribution, the gradient for a batch will not be zero but will have
components in the direction of shifting and scaling the codes to reduce
the MMD for a given batch. Code normalization directly takes care
of this reduction, and allows the training process to spend its effort
on improving the reconstruction error. Technically, this is achieved
by projecting out the components of the gradient corresponding to
shifting and scaling which is automatically achieved by normalization,
see \cite[Section 3, penultimate paragraph]{BatchRenorm}.

This observation reveals an interesting aspect of training with the
MMD as compared to training in an adversarial manner \cite{AdversAE}.
When training in an adversarial manner, the goal is to make the codes
in each batch resemble a sample from the standard multivariate normal
distribution. At an intuitive level, we expect this would happen with
the MMD penalty as well. However, this is not the case---we see that,
on average, the MMD penalty considers normalized samples more ``ideal''
than the actual samples from the target distribution. Luckily, the
neural network cannot learn batch-wise operations (e.g. it cannot
learn to do \emph{batch-wise} normalization or whitening by \emph{itself})
assuming that at inference time the inputs are processed independently
of each other. As a result, this phenomenon will not prevent convergence
to the target distribution. A rigorous argument follows from unbiasedness,
$\mathbb{E}[\mathrm{MMD}_{u}^{2}]=\mathrm{MMD}^{2}\geq0$ where the
expectation is taken over i.i.d. samples and equality holds only at
convergence to the target distribution; this makes any overall shift
to the left at the inference time impossible.

\emph{Remark:} When monitoring neural net training the following should
be taken into account to avoid wrongly declaring that overfitting
has occurred. When code normalization is used, the distribution shift
exemplified in Figure \ref{fig:codenorm-effect} will result in a
noticeably smaller training loss than the validation loss. This is
because code normalization uses the batch statistics during training
and population statistics at validation/test time. The difference
between these losses can be on the order of several $\lambda$'s;
here $\lambda$ is the regularization strength. This effect is akin
to substituting an estimator of a parameter (computed from the same
data) into a statistic and results in a distributional changes. 

\paragraph*{Avoiding Outliers}

Another benefit of code normalization is that it provides a solution
to outlier insensitivity problem of the MMD penalty, described below.
Indeed, scaling by the standard deviation (rather than by a robust
surrogate) controls the tail behavior of the code distribution. Due
to this control, the code distribution ends up having a light tail
and no code falls too far away from the origin. 

The outlier insensitivity problem is not specific to our closed-form
formula or the choice of the kernel (see Section \ref{sec:Experiments}
for an empirical verification); this problem is relevant to any kernel
$k(x,y)=f(\Vert x-y\Vert)$ such that $f(r)\rightarrow0$ as $r\rightarrow\infty$.
Given a sample $Q_{n}=\{z_{i}\}_{i=1}^{n}$ from the standard multi-variate
normal distribution, consider a modified sample $Q'_{n}=\{z'_{i}\}_{i=1}^{n}$,
where $z'_{i}=z_{i},i=2,...,n$ and $z'_{1}$ is far from the origin.
Expressing the sum of vanishingly small exponentials via the $O$-notation,
we can compute the difference in MMD incurred by this change:
\begin{align}
\Delta\mathrm{MMD}_{u}^{2} & =\mathrm{MMD}_{u}^{2}(Q_{n}')-\mathrm{MMD}_{u}^{2}(Q_{n})=\label{eq:mmd_delta}\\
 & =\frac{2}{n}\left[\left(\frac{\gamma^{2}}{1+\gamma^{2}}\right)^{d/2}e^{-\frac{\Vert z_{1}\Vert^{2}}{2(1+\gamma^{2})}}-\frac{1}{n-1}\sum_{j\neq1}^{n}e^{-\frac{\Vert z_{1}-z_{j}\Vert^{2}}{2\gamma^{2}}}+O\left(e^{-\frac{\Vert z'_{1}\Vert^{2}}{2(1+\gamma^{2})}}\right)\right]\nonumber 
\end{align}
Note that the second term is the sample average approximation of $\mathbb{E}_{x\sim Q=P}[e^{-\Vert z_{1}-x\Vert^{2}/(2\gamma^{2})}]$.
This expectation can be computed analytically (in fact it is equivalent
to the summand in the second term of Eq. (\ref{eq:mmdu})) and it
precisely cancels the first term here in Eq. (\ref{eq:mmd_delta}),
giving $\mathbb{E}[\Delta\mathrm{MMD}_{u}^{2}]\approx0$. Thus, MMD
changes very little despite the presence of the large outlier.

Given the mixed objective and stochasticity inherent in the training
process, this issue has an effect on WAE training even before reaching
the limits of computer precision. Indeed, in addition to the MMD penalty,
the WAE objective contains the reconstruction term. Given the incentive
to reconstruct well, the optimizer will realize that it is beneficial
to push some of the codes far away from the origin, since the origin
is where most of the codes concentrate. If this happens only for a
few codes in a batch, the MMD penalty will not be big enough so as
to pull these codes back towards the origin. As a result, the training
process will result in a distribution $Q$ that has outliers. Our
experiments show that the proposed code normalization provides a solution
to this issue without a need for using adaptive kernel widths or extra
penalties.

\section{\label{sec:Experiments}Experiments}

First we discuss our parameterization for the kernel width used in
computation of various MMD measures. A rule of thumb choice of the
kernel width is $\gamma^{2}=d$, where $d$ is the dimension of the
code space (see e.g. \cite{InfoVAE,WAE}). This choice is based on
considering the average pair-wise distance between two points drawn
from the standard multi-variate normal distribution, and halving it
to offset the multiplication by $2$ in the expression for the kernel.
We will see that this choice gives rather suboptimal results, yet
it provides a good point of reference for defining scale of the kernel
as $s=\gamma^{2}/d$. We will experiment with various choices of $s$,
where $s>1$ gives wider and $s<1$ gives narrower kernels. The sample
size in the experiments is chosen as $n=100$ in agreement with a
commonly used batch size while training neural networks.

\paragraph*{Validation}

We first experimentally verify that our closed-form formula for SMMD
results in zero mean and unit variance under the null when $Q=\mathcal{N}_{d}$.
To this end, we sample $n=100$ points from the standard $d$-variate
normal distribution and compute the value of $\mathrm{SMMD}^{2}$.
This process is repeated 10,000 times to obtain the empirical distribution
of the values. Figure \ref{fig:Variance-curve} shows the violin plots
of these empirical distributions computed for several values of the
kernel scale $s$ and dimensionality $d$. The red segments in this
plot are centered at the mean, and they extend between mean $\pm$
standard deviation. We observe from the graph that the means are close
to zero and the standard deviations are close to 1 as expected.

\begin{figure}
\begin{centering}
\includegraphics[width=1\textwidth]{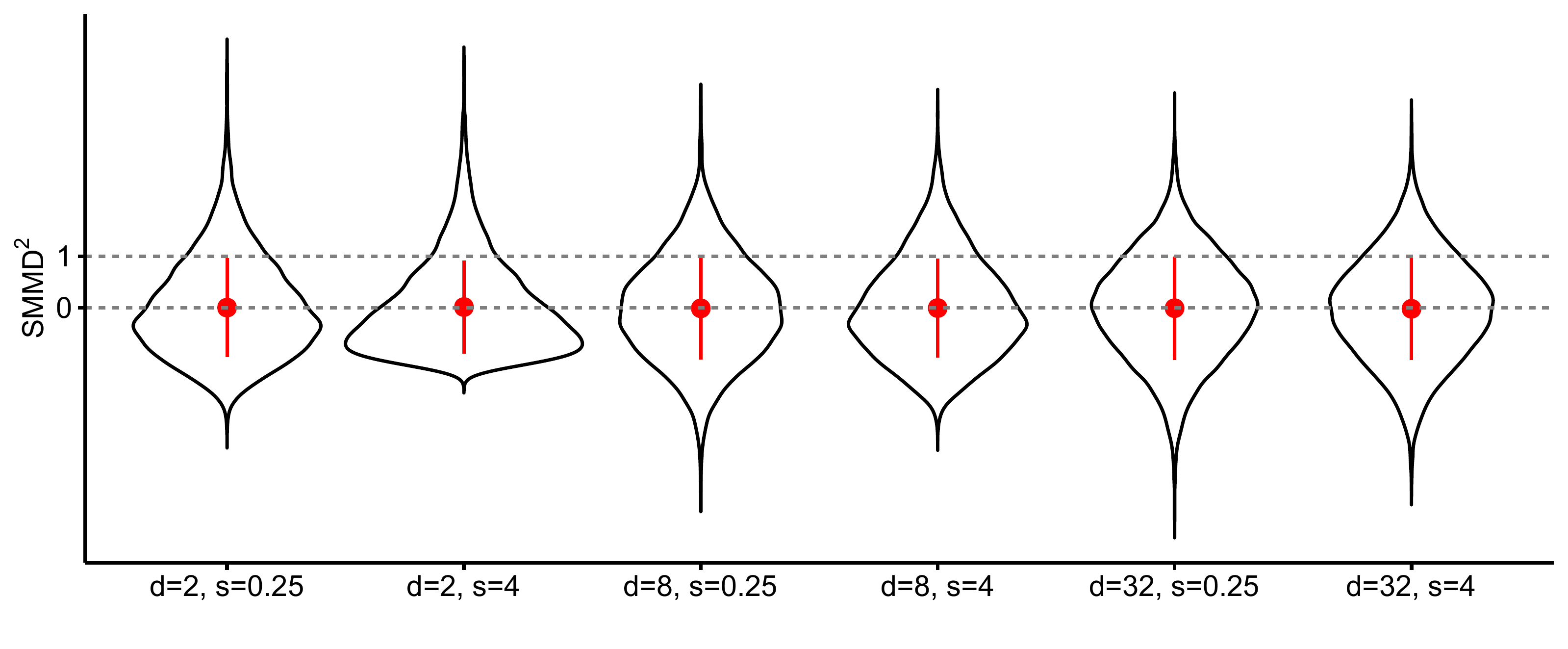}
\par\end{centering}
\caption{\label{fig:violin_plot}Violin plots verify that $\mathrm{\mathrm{SMMD}^{2}}$
has zero mean and unit variance under the null. Here, batch size is
$n=100$ and the kernel width is expressed via the scale $s$ as $\gamma^{2}=s\cdot d$}
\end{figure}

\paragraph*{Discriminative Performance}

The goal of the next experiment is to compare our closed-formula estimator
of MMD (referred to as ``Analytic RBF'') to the standard sampling
based estimator using the same Gaussian RBF kernel (``Empirical RBF'').
We also compare to the sampling based estimator but with the inverse
multi-quadratics (IMQ) kernel defined by $k(x,y)=1/\left(1+\Vert x-y\Vert^{2}/(2\gamma^{2})\right)$;
we call this ``Empirical IMQ''. The IMQ kernel is often claimed
to be superior to the RBF kernel due to its slower tail decay.

In our first experiment we would like to determine which one of these
three methods is most effective at distinguishing the standard $d$-variate
normal distribution from the uniform distribution. Since our goal
is to train neural networks rather than perform hypothesis testing,
we will not use the test power as a metric of interest; instead we
will rely on the effect size defined below. In addition, we are not
studying the dependence on the latent dimension, so we do not have
to worry about the fair choice of alternatives \cite{Ramdas_decreasing_power}.

\begin{figure}
\begin{centering}
\begin{tabular}{lccccccc}
\toprule 
Method & Kernel Scale & $d=1$ & $d=2$ & $d=4$ & $d=8$ & $d=16$ & $d=32$\tabularnewline
\midrule
\midrule 
an RBF & 2 & $0.49$ & $0.14$ & $0.05$ & $0.02$ & $3.7e-04$ & $0.11$\tabularnewline
 & 1 & $1.18$ & $0.7$ & $0.29$ & $0.09$ & $0.04$ & $0.03$\tabularnewline
 & 1/2 & $1.85$ & $1.64$ & $0.97$ & $0.45$ & $0.17$ & $0.03$\tabularnewline
 & 1/4 & $1.97$ & $2.32$ & $2.02$ & $1.19$ & $0.76$ & $0.35$\tabularnewline
 & 1/8 & $\mathbf{2.28}$ & $\mathbf{2.61}$ & $\mathbf{2.56}$ & $2.01$ & $\mathbf{1.5}$ & $0.98$\tabularnewline
 & 1/16 & $\mathbf{2.21}$ & $\mathbf{2.62}$ & $\mathbf{2.49}$ & $1.88$ & $1.39$ & $\mathbf{1.17}$\tabularnewline
 & 1/32 & $\mathbf{2.22}$ & $\mathbf{2.49}$ & $2.1$ & $1.16$ & $0.51$ & $0.44$\tabularnewline
 & HZ & $1.46$ & $1.99$ & $2.26$ & $\mathbf{2.13}$ & $\mathbf{1.49}$ & $0.39$\tabularnewline
\midrule 
emp RBF & 2 & $0.32$ & $0.15$ & $0.15$ & $9.1e-04$ & $0.03$ & $0.13$\tabularnewline
 & 1 & $0.75$ & $0.42$ & $0.16$ & $0.01$ & $0.02$ & $0.01$\tabularnewline
 & 1/2 & $1.07$ & $0.95$ & $0.5$ & $0.23$ & $0.09$ & $0.01$\tabularnewline
 & 1/4 & $1.36$ & $1.34$ & $1.1$ & $0.66$ & $0.3$ & $0.14$\tabularnewline
 & 1/8 & $1.35$ & $\mathbf{1.57}$ & $1.3$ & $\mathbf{0.99}$ & $0.66$ & $0.51$\tabularnewline
 & 1/16 & $\mathbf{1.4}$ & $\mathbf{1.54}$ & $\mathbf{1.38}$ & $\mathbf{1.02}$ & $\mathbf{0.71}$ & $\mathbf{0.62}$\tabularnewline
 & 1/32 & $\mathbf{1.45}$ & $1.36$ & $1.1$ & $0.57$ & $0.26$ & $0.15$\tabularnewline
 & HZ & $1.07$ & $1.18$ & $1.23$ & $\mathbf{0.98}$ & $0.64$ & $0.25$\tabularnewline
\midrule 
emp IMQ & 2 & $0.51$ & $0.29$ & $0.08$ & $0.06$ & $0.01$ & $0.03$\tabularnewline
 & 1 & $0.74$ & $0.48$ & $0.27$ & $0.12$ & $0.06$ & $0.03$\tabularnewline
 & 1/2 & $1.01$ & $0.88$ & $0.49$ & $0.33$ & $0.08$ & $0.04$\tabularnewline
 & 1/4 & $1.21$ & $1.12$ & $0.74$ & $0.46$ & $0.16$ & $0.06$\tabularnewline
 & 1/8 & $1.23$ & $1.36$ & $1.1$ & $0.45$ & $0.2$ & $0.1$\tabularnewline
 & 1/16 & $\mathbf{1.32}$ & $\mathbf{1.41}$ & $\mathbf{1.22}$ & $0.53$ & $0.27$ & $0.07$\tabularnewline
 & 1/32 & $\mathbf{1.31}$ & $\mathbf{1.46}$ & $1.19$ & $0.61$ & $\mathbf{0.32}$ & $0.08$\tabularnewline
 & 1/64 & $\mathbf{1.33}$ & $1.38$ & $1.17$ & $0.63$ & $0.29$ & $\mathbf{0.2}$\tabularnewline
 & 1/128 & $\mathbf{1.33}$ & $\mathbf{1.42}$ & $\mathbf{1.26}$ & $\mathbf{0.82}$ & $\mathbf{0.32}$ & $0.08$\tabularnewline
 & 1/256 & $\mathbf{1.32}$ & $1.26$ & $1.19$ & $0.63$ & $0.28$ & $0.13$\tabularnewline
 & 1/512 & $1.16$ & $1.04$ & $1.14$ & $0.77$ & $0.27$ & $0.08$\tabularnewline
 & 1/1024 & $1.08$ & $1.04$ & $1.06$ & $0.77$ & $\mathbf{0.32}$ & $0.11$\tabularnewline
\bottomrule
\end{tabular}
\par\end{centering}
\captionof{table}{\label{tab:Discrimination-power-table-unifrom}Discrimination power between $d$-variate standard normal distribution and uniform $U[-\sqrt{3},\sqrt{3}]^{d}$ distribution as measured by $\tau$.}

\vfill{}

\begin{centering}
\includegraphics[width=1\textwidth]{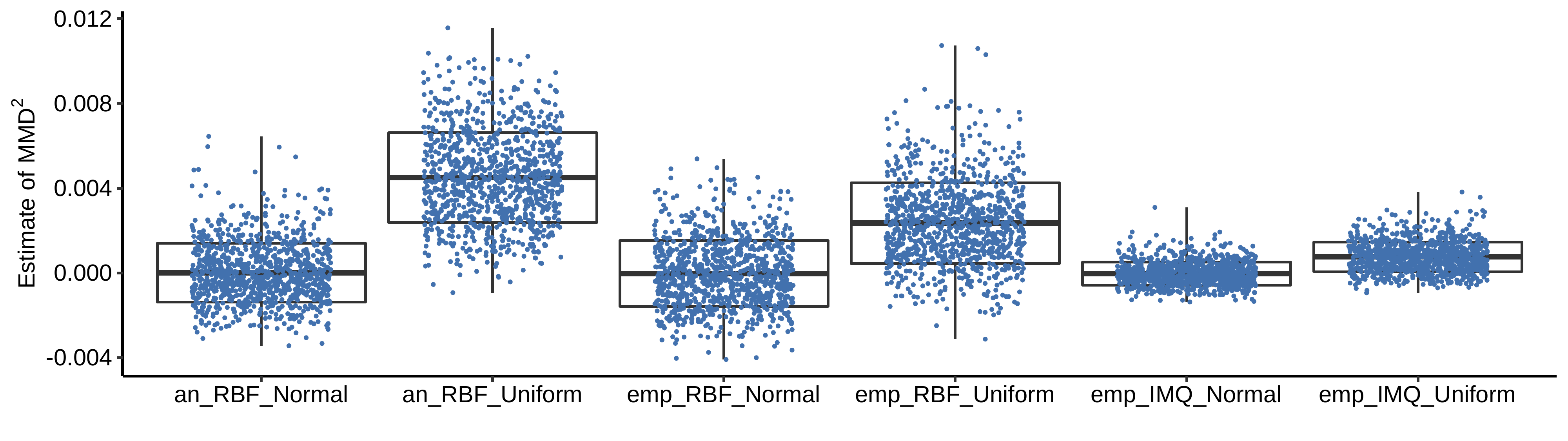}
\par\end{centering}
\captionof{figure}{\label{fig:boxplot-discrimination}Graphical representation of this discrimination experiment for $d=8$. }
\end{figure}

\begin{figure}
\begin{centering}
\begin{tabular}{lccccccc}
\toprule 
Method & Kernel Scale & $d=1$ & $d=2$ & $d=4$ & $d=8$ & $d=16$ & $d=32$\tabularnewline
\midrule
\midrule 
an RBF & 2 & $0.17$ & $0.12$ & $0.11$ & $0.11$ & $0.1$ & $0.1$\tabularnewline
 & 1 & $0.35$ & $0.33$ & $0.43$ & $0.5$ & $0.39$ & $0.31$\tabularnewline
 & 1/2 & $0.59$ & $0.69$ & $1.61$ & $1.49$ & $1.46$ & $1.23$\tabularnewline
 & 1/4 & $0.86$ & $1.14$ & $3.07$ & $3.64$ & $3.75$ & $3.11$\tabularnewline
 & 1/8 & $1.1$ & $1.42$ & $4.6$ & $\mathbf{4.95}$ & $\mathbf{5.37}$ & $\mathbf{4.07}$\tabularnewline
 & 1/16 & $1.18$ & $1.9$ & $\mathbf{4.86}$ & $4.71$ & $4.24$ & $3.21$\tabularnewline
 & 1/32 & $\mathbf{1.34}$ & $\mathbf{2.1}$ & $4.25$ & $3.37$ & $2.75$ & $2.25$\tabularnewline
 & HZ & $0.52$ & $1$ & $3.92$ & $\mathbf{5.19}$ & $4.18$ & $2.5$\tabularnewline
\midrule 
emp RBF & 2 & $0.03$ & $0.14$ & $0.05$ & $0.07$ & $0.06$ & $0.03$\tabularnewline
 & 1 & $0.11$ & $0.09$ & $0.32$ & $0.24$ & $0.19$ & $0.19$\tabularnewline
 & 1/2 & $0.24$ & $0.46$ & $0.88$ & $0.81$ & $0.72$ & $0.59$\tabularnewline
 & 1/4 & $0.53$ & $0.67$ & $1.69$ & $1.98$ & $2.42$ & $2.16$\tabularnewline
 & 1/8 & $0.6$ & $0.77$ & $2.52$ & $3.36$ & $\mathbf{4.22}$ & $\mathbf{3.68}$\tabularnewline
 & 1/16 & $0.73$ & $0.98$ & $\mathbf{2.68}$ & $\mathbf{3.74}$ & $3.81$ & $3.35$\tabularnewline
 & 1/32 & $\mathbf{0.78}$ & $\mathbf{1.16}$ & $2.53$ & $3$ & $2.69$ & $2.11$\tabularnewline
 & HZ & $0.26$ & $0.55$ & $2.05$ & $\mathbf{3.58}$ & $3.86$ & $2.41$\tabularnewline
\midrule 
emp IMQ & 2 & $4.7e-03$ & $0.19$ & $0.25$ & $0.16$ & $0.12$ & $0.16$\tabularnewline
 & 1 & $0.16$ & $0.26$ & $0.49$ & $0.33$ & $0.43$ & $0.34$\tabularnewline
 & 1/2 & $0.31$ & $0.29$ & $0.78$ & $0.78$ & $0.68$ & $0.62$\tabularnewline
 & 1/4 & $0.4$ & $0.49$ & $1.21$ & $1.26$ & $1.25$ & $1.06$\tabularnewline
 & 1/8 & $0.51$ & $0.73$ & $1.78$ & $1.85$ & $1.73$ & $1.47$\tabularnewline
 & 1/16 & $0.54$ & $0.82$ & $2.06$ & $2.17$ & $2.14$ & $1.89$\tabularnewline
 & 1/32 & $0.75$ & $1.08$ & $2.18$ & $2.49$ & $2.32$ & $1.97$\tabularnewline
 & 1/64 & $0.92$ & $\mathbf{1.18}$ & $2.3$ & $2.7$ & $2.43$ & $1.95$\tabularnewline
 & 1/128 & $0.89$ & $\mathbf{1.24}$ & $\mathbf{2.49}$ & $2.69$ & $\mathbf{2.57}$ & $2.11$\tabularnewline
 & 1/256 & $0.93$ & $\mathbf{1.21}$ & $\mathbf{2.46}$ & $\mathbf{2.82}$ & $\mathbf{2.54}$ & $\mathbf{2.22}$\tabularnewline
 & 1/512 & $\mathbf{1.11}$ & $\mathbf{1.22}$ & $2.32$ & $\mathbf{2.88}$ & $\mathbf{2.58}$ & $\mathbf{2.26}$\tabularnewline
 & 1/1024 & $\mathbf{1.05}$ & $1.15$ & $2.23$ & $\mathbf{2.77}$ & $\mathbf{2.61}$ & $\mathbf{2.16}$\tabularnewline
\bottomrule
\end{tabular}
\par\end{centering}
\captionof{table}{\label{tab:Discrimination-power-table-mnist}Discrimination power between the $d$-variate standard normal distribution and a latent $d$-dimensional embedding of MNIST with an unregularized autoencoder.}

\vfill{}

\begin{centering}
\includegraphics[width=1\textwidth]{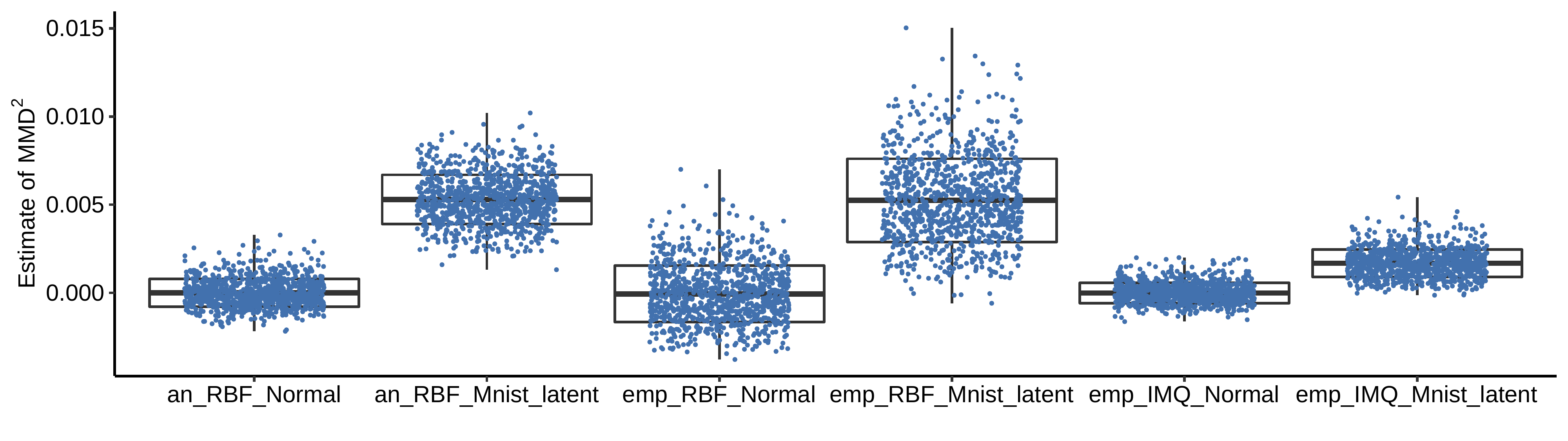}
\par\end{centering}
\captionof{figure}{\label{fig:boxplot-discrimination-mnist}Graphical representation of the discrimination experiment with MNIST latent embedding for $d=8$.}
\end{figure}

The uniform distribution under consideration is $U[-\sqrt{3},\sqrt{3}]^{d}$.
Note that this particular uniform distribution has mean 0 and variance
1 in each dimension just like the normal distribution; distinguishing
the two distributions requires going beyond the first two moments.
For each of the three methods, for a fixed dimension $d$ and kernel
scale $s$, we sample $n=100$ points from the the standard $d$-variate
normal distribution and compute the corresponding MMD estimate. Next
we sample $n=100$ points from the uniform distribution $U[-\sqrt{3},\sqrt{3}]^{d}$
and compute the corresponding MMD estimate. We repeat this 200 times,
and compute the corresponding means $\mathrm{Mean}_{1}$ and $\mathrm{Mean}_{2}$,
and the standard deviations $\mathrm{SD}_{1}$ and $\mathrm{SD}_{2}$
corresponding to each of the two sets of 200 MMD values\footnote{Of course, we expect $\mathrm{Mean}_{1}\approx0$ since all of the
three methods are unbiased. For the Analytic RBF, we also know the
theoretical value of $\mathrm{SD}_{1}$ from the closed-form formula
for the variance. However, for fairness we will use empirical estimates
for all of the three methods.}. Now we can measure the discriminativeness of a given method by computing
$\tau(\mathrm{method},s,d)=\frac{\mathrm{\vert Mean}_{1}-\mathrm{Mean}_{2}\vert}{(\mathrm{SD}_{1}+\mathrm{SD}_{2})/2}.$
Note that this is the effect size of a two sample t-test as measured
by Cohen's d \cite{cohen1988spa}. Larger values of $\tau$ mean better
discrimination, which potentially translates to better gradients for
neural network training.

The results are presented in Table \ref{tab:Discrimination-power-table-unifrom}.
Note that the experiment was done for different values of the kernel
scale; due to the heavier tail, we included more scale choices for
the IMQ kernel than for the RBF kernel. For each method and dimensionality
choice $d$, the best performing choice of the kernel scale corresponds
to the maximum value of $\tau$; these $\tau$ values are shown in
boldface (we also highlight the $\tau$ values that are within $5\%$
of the maximum). Figure \ref{fig:boxplot-discrimination} provides
a box-plot display (whiskers span the range of all of the values)
for this experiment when $d=8$. In this graph, for each method, the
best choice of the kernel scale was used; when the boxes corresponding
to normal and uniform distributions overlap, it means that the method
has difficulty discriminating the two distributions. In terms of training
neural networks, this means that the corresponding MMD penalty may
not be able to provide a strong gradient direction for training because
the difference is lost within the stochastic noise.

We repeat the same experiment but instead of the uniform distribution
we use a distribution obtained from a neural networks. We use the
MNIST dataset and train auto-encoders (both encoder and decoder have
two hidden layers with 128 neurons each, ReLU activations) with different
latent dimensions $d$ with no regularization. The codes corresponding
to the test data are extracted and shifted to have zero mean. We observed
that with growing $d$ the various latent dimensions were highly correlated
(e.g. Pearson correlations as high as $0.4$); thus, to make the task
more difficult, we applied PCA-whitening to the latent codes. The
resulting discrimination performance is presented in Table \ref{tab:Discrimination-power-table-mnist}
and Figure \ref{fig:boxplot-discrimination-mnist}. 

By examining both of the tables above, we can see that Analytic RBF
method outperforms both the Empirical RBF and IMQ methods in terms
of discrimination power. Another observation is that the commonly
recommended choice of $\gamma^{2}=d$ (which corresponds to the kernel
scale $s=1$) is never a good choice; a similar finding for the median
heuristic was spelled out in \cite{Sutherland_model_criticismMMD}.
The kernel width recommended for Henze-Zirkler test gives mixed results,
which is somewhat expected---optimality for density estimation does
not guarantee optimal discriminative performance. Examining the Analytic
RBF results, it seems that kernel scales $s=1/8$ or $s=1/16$ provide
a good rule of thumb choices. Finally, in these particular examples
we see that despite its having a larger repertoire of kernel scale
choices, Empirical IMQ does not perform as well as Empirical RBF.
While these results are limited to two datasets, yet they bring into
question the commonly recommended choices of the kernel and its width.
Of course, our analysis assumes that the alternative distribution
has zero mean and unit variance in each dimension. We believe that
this is the most relevant setting to WAE learning because during the
late stages of WAE training the code distribution starts converging
to the normal distribution. 

\paragraph*{Outliers}

\begin{figure}
\begin{centering}
\includegraphics[width=1\textwidth]{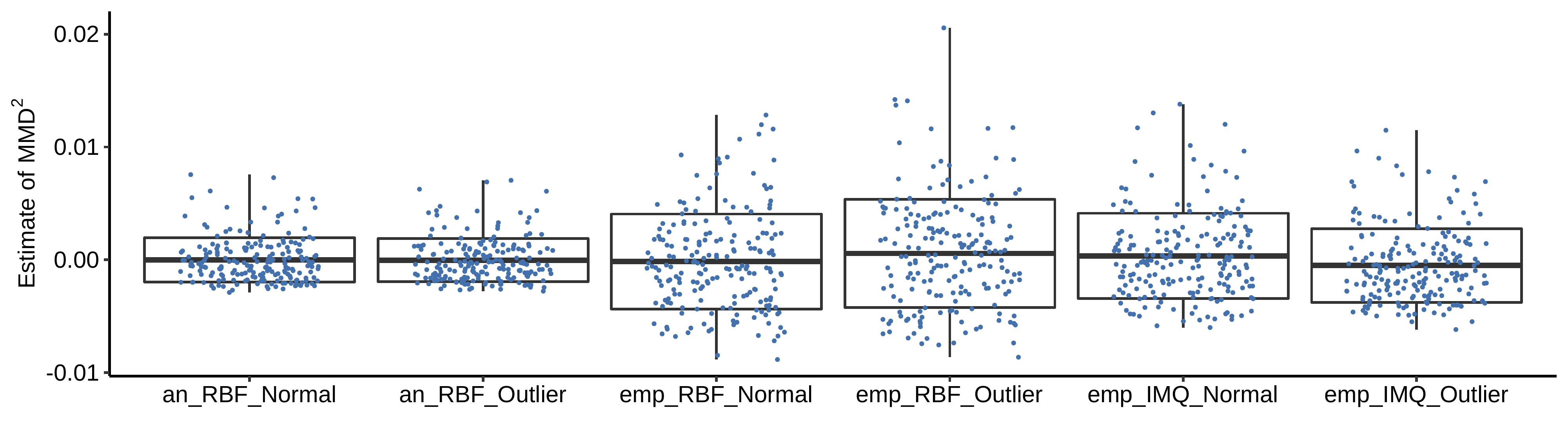}
\par\end{centering}
\caption{\label{fig:Outlier-discrimination-experimen}Outlier discrimination
experiment carried out for $d=4$. For each method, the most discriminative
(i.e. maximum $\tau$) kernel scale is chosen.}
\end{figure}

Here we experimentally verify the outlier insensitivity of the MMD
and demonstrate that the issue is not peculiar to our approach. To
this end, we run the discrimination experiment above but this time
trying to distinguish a sample from the standard $d$-variate normal
distribution from the same but with one of the sample points replaced
with a point far away from the origin (namely $z_{1}\rightarrow z'_{1}=100\cdot\vec{1}$).
Figure \ref{fig:Outlier-discrimination-experimen} shows that all
of the three methods fail to distinguish these two distributions in
practice.

\begin{figure}
\begin{centering}
\setlength{\tabcolsep}{3pt}%
\begin{tabular}{cccc}
\rotatebox{90}{\qquad Code Norm (60 epochs)} & \includegraphics[width=0.31\textwidth]{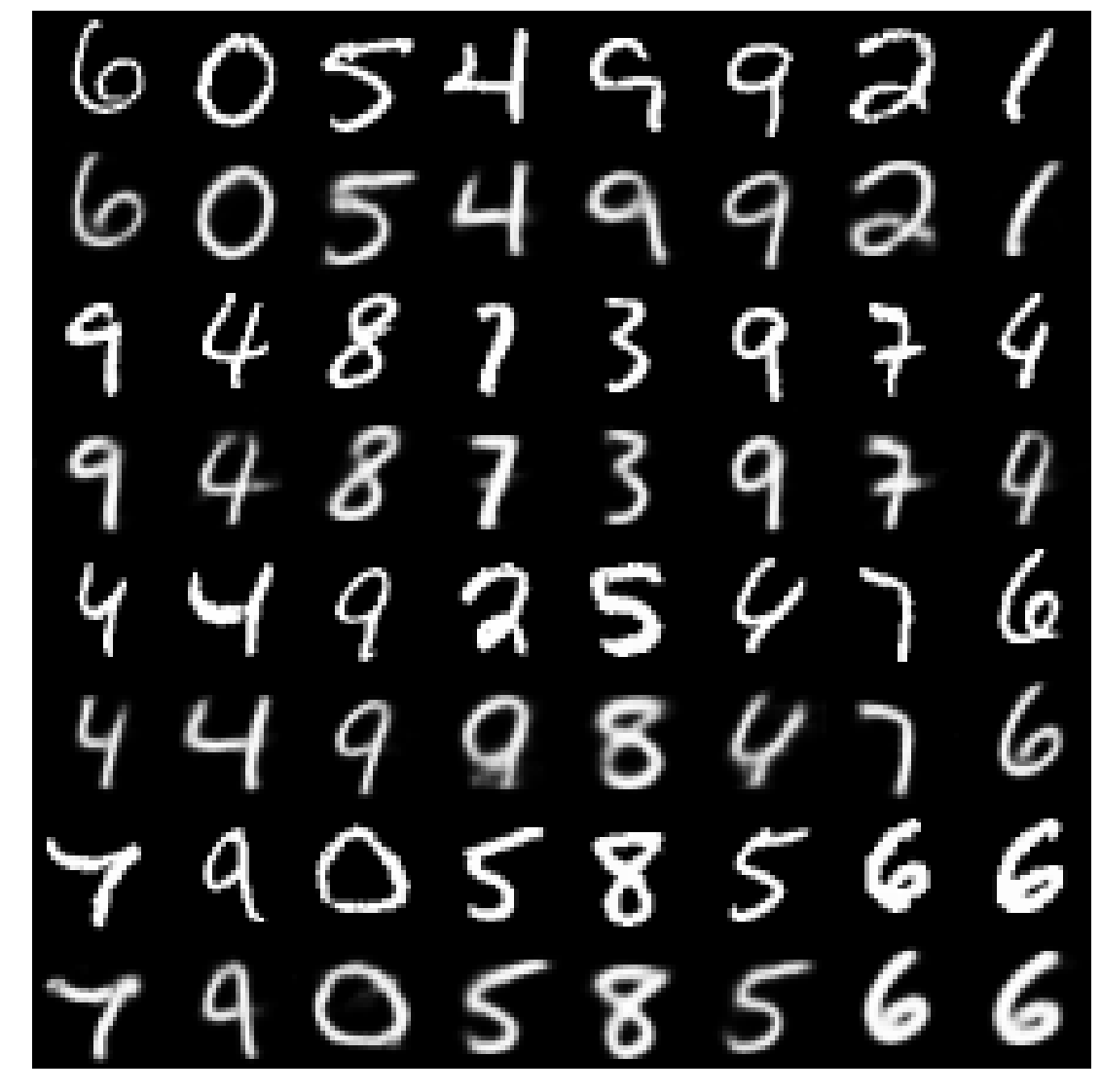} & \includegraphics[width=0.31\textwidth]{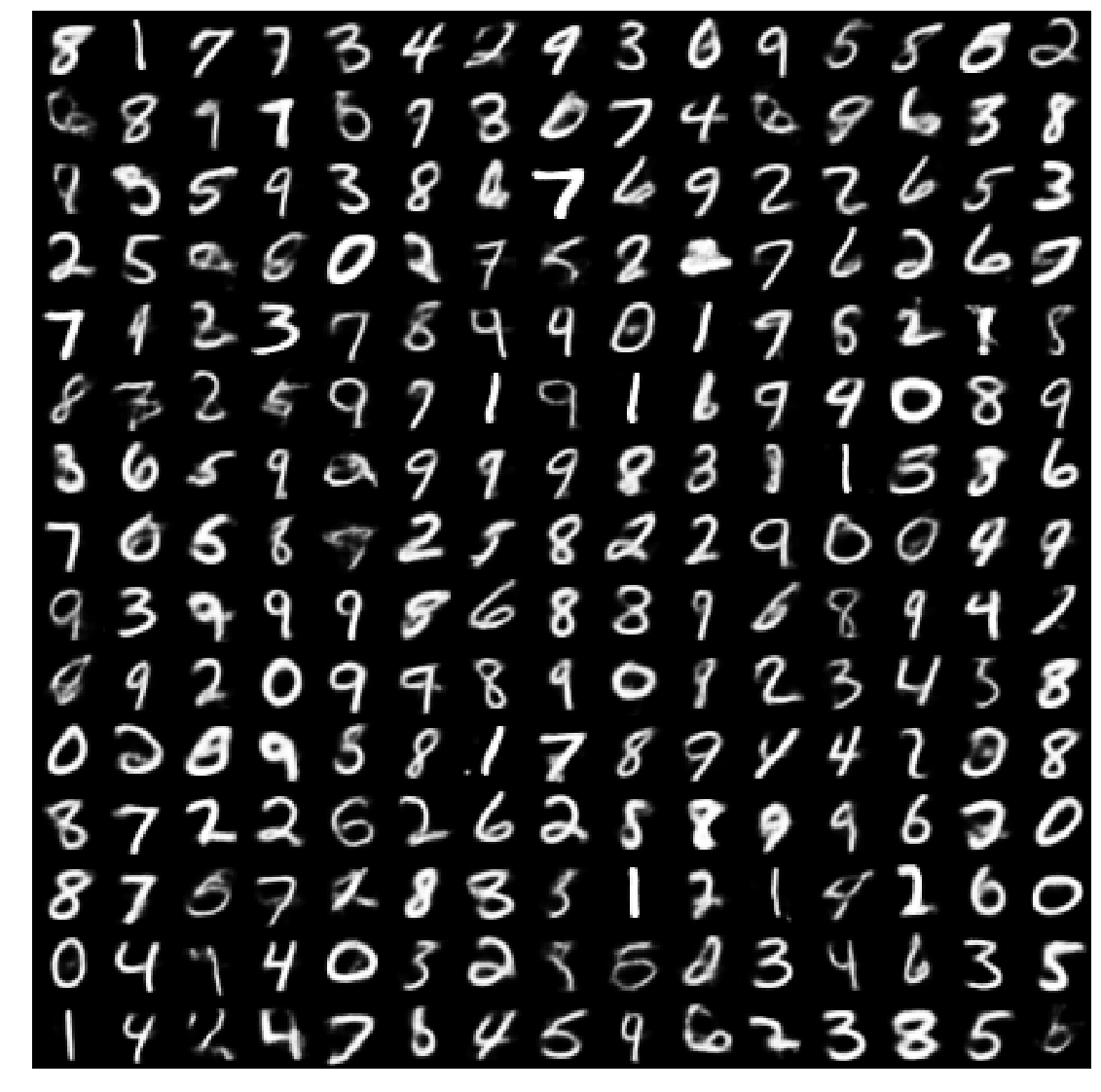} & \includegraphics[width=0.31\textwidth]{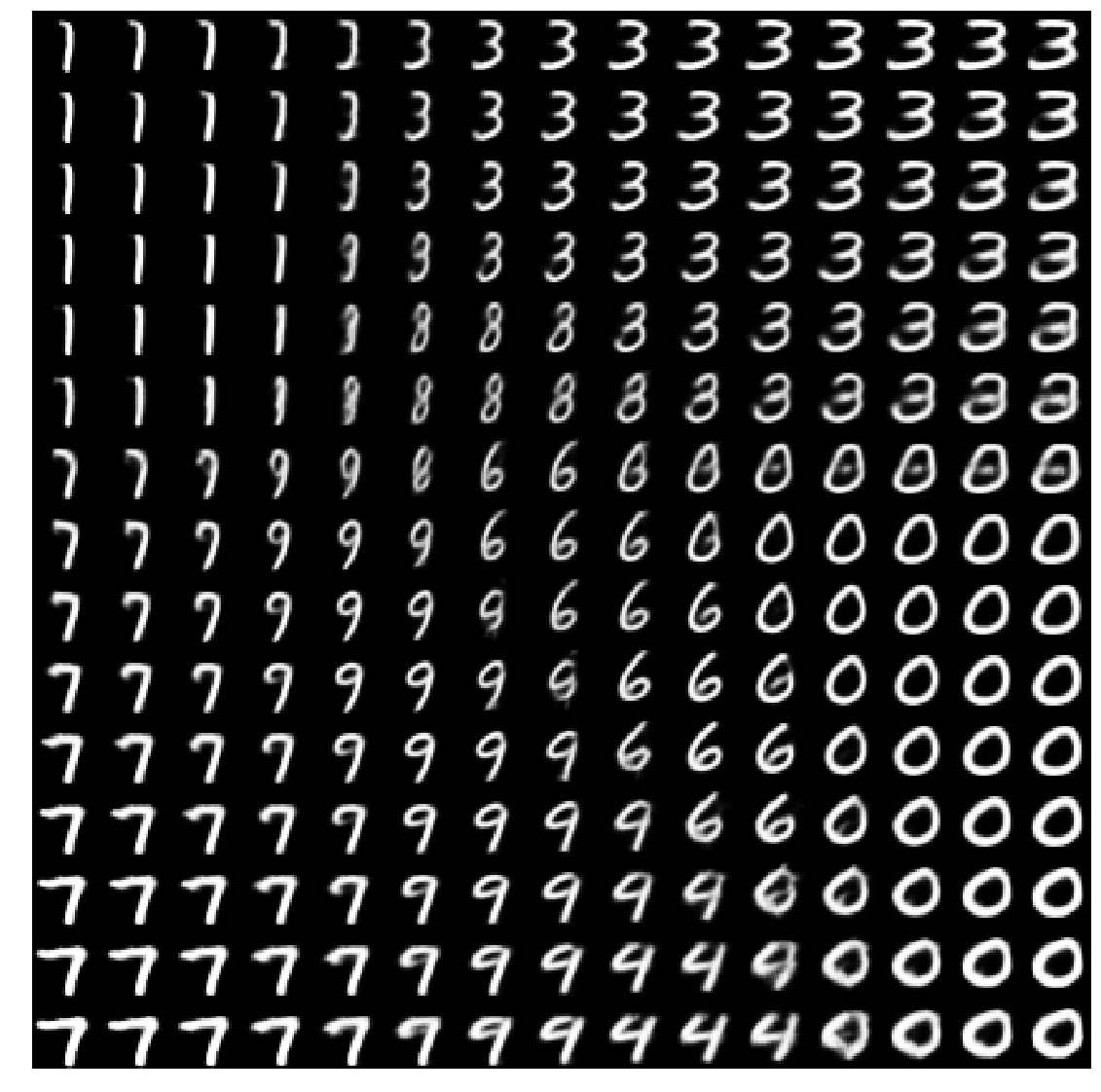}\tabularnewline
\rotatebox{90}{\qquad AdaptiveBN (80 epochs)} & \includegraphics[width=0.31\textwidth]{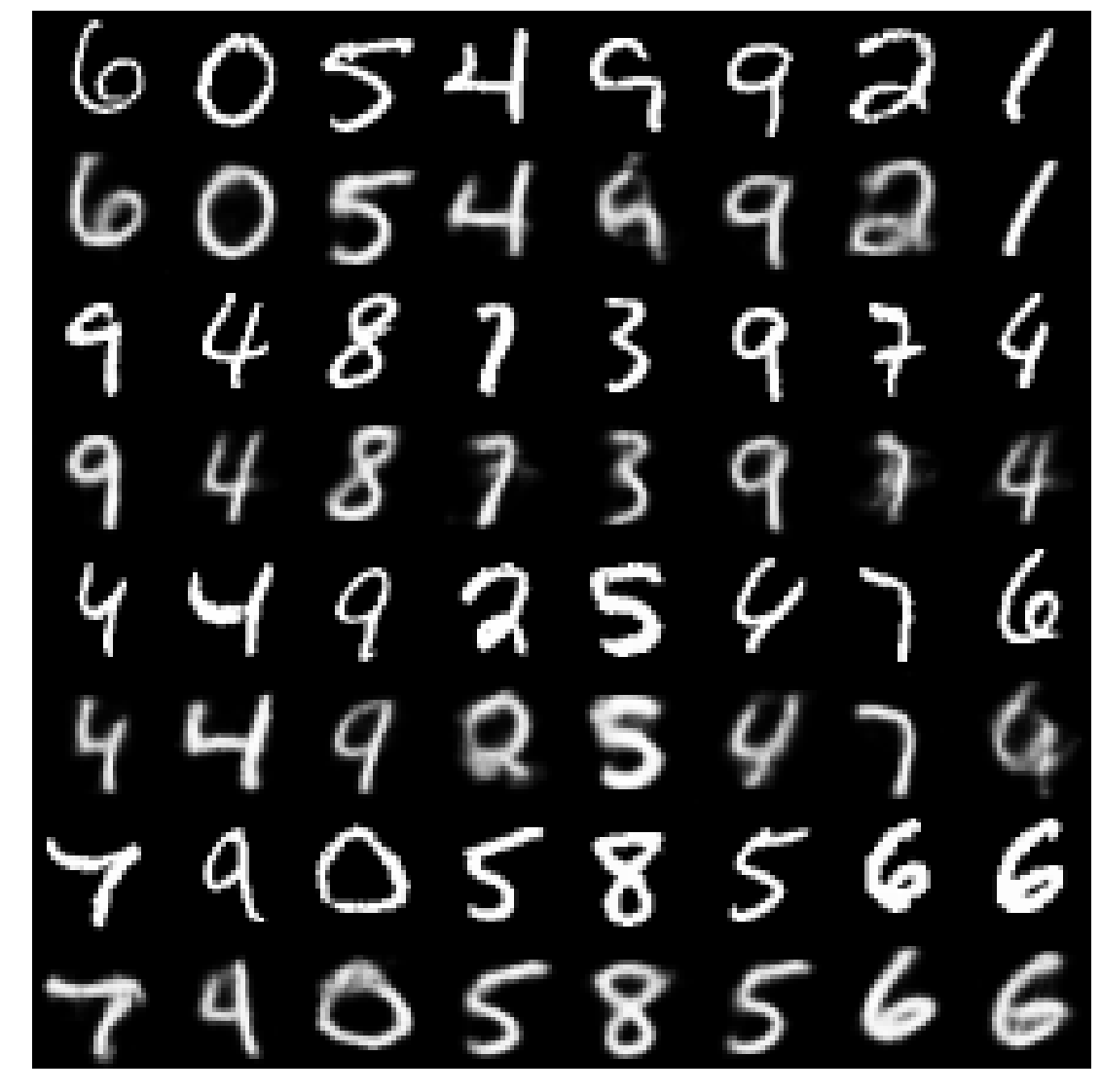} & \includegraphics[width=0.31\textwidth]{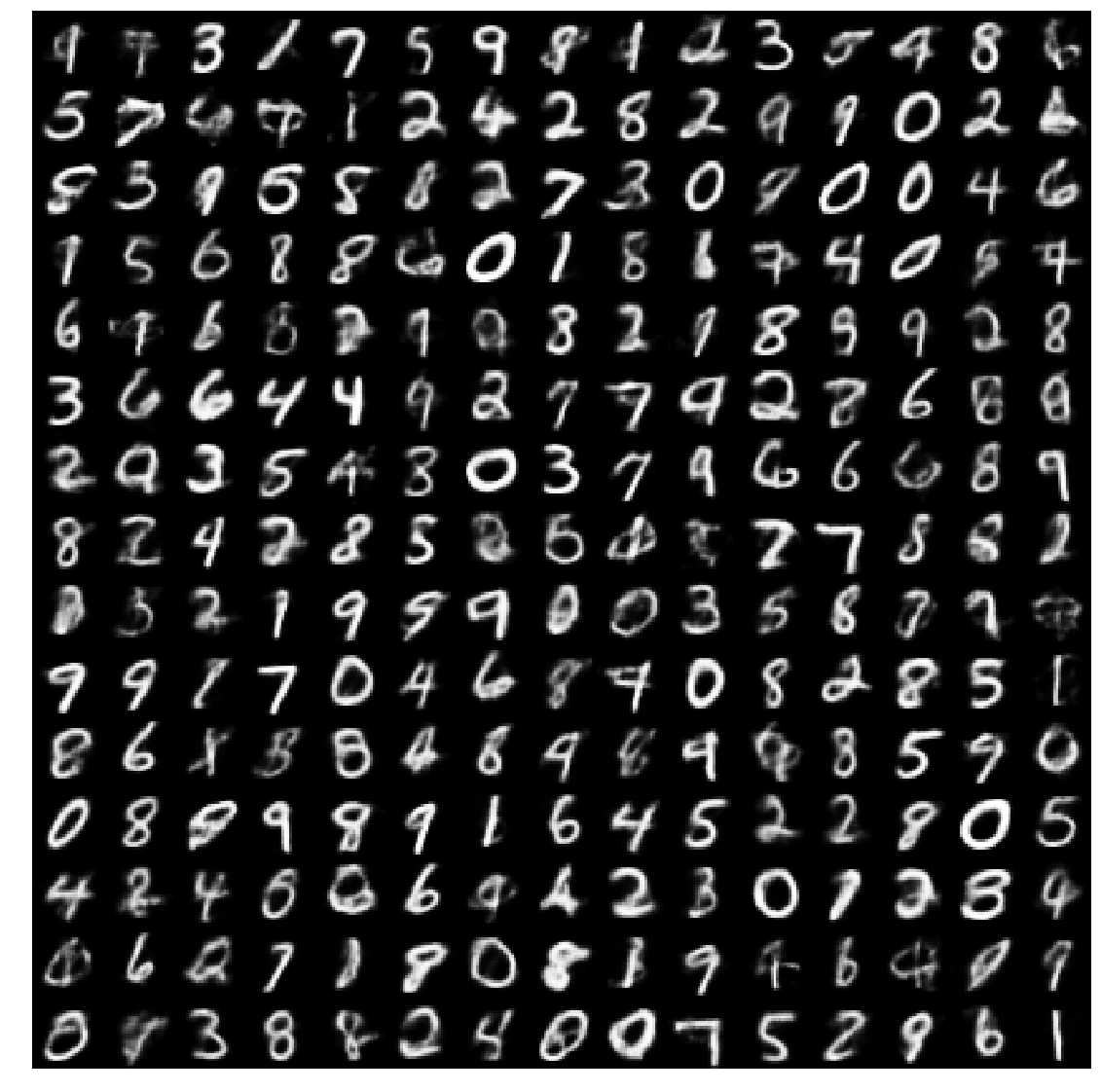} & \includegraphics[width=0.31\textwidth]{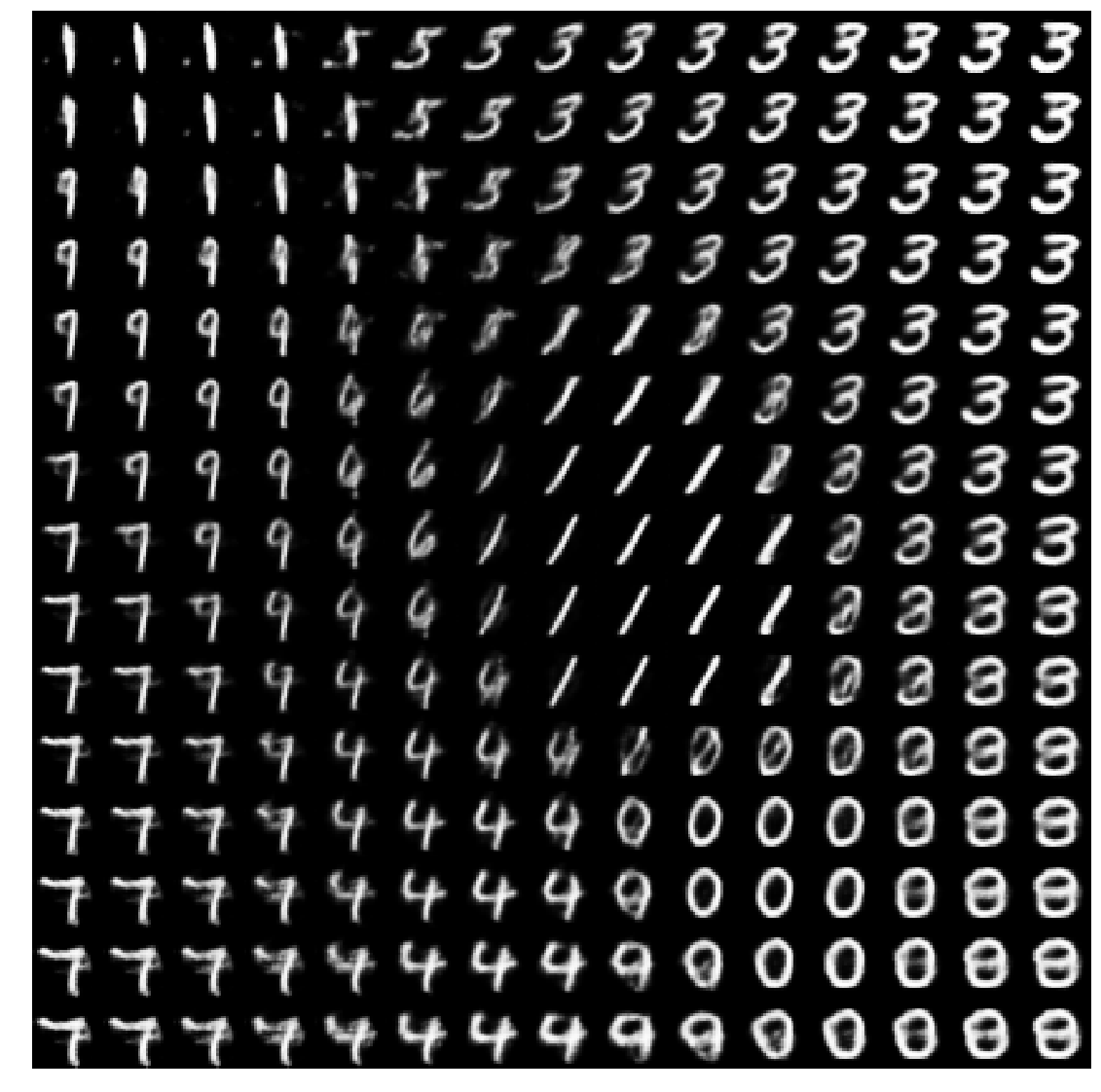}\tabularnewline
\rotatebox{90}{\qquad AdaptivePlain (80 epochs)} & \includegraphics[width=0.31\textwidth]{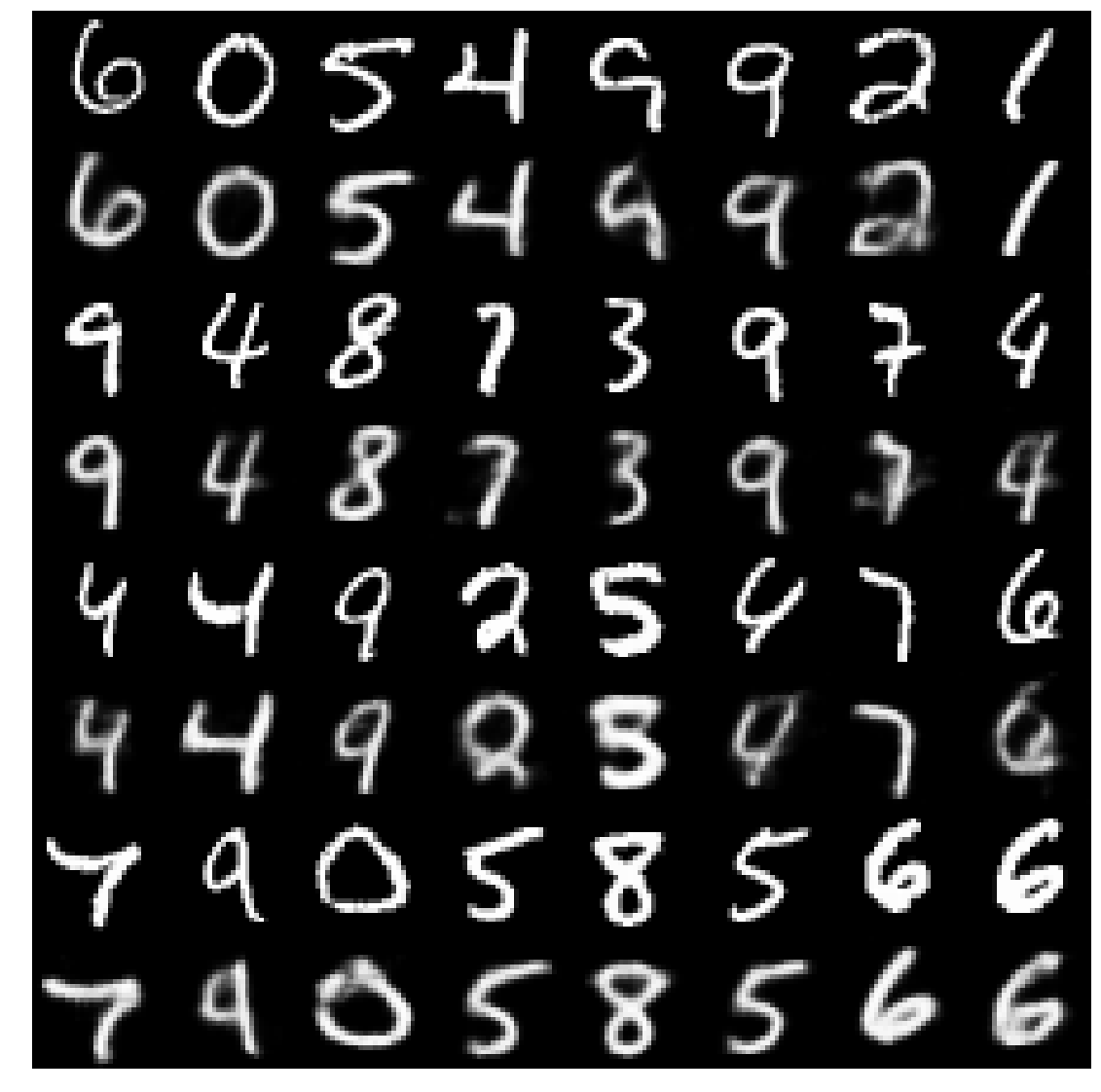} & \includegraphics[width=0.31\textwidth]{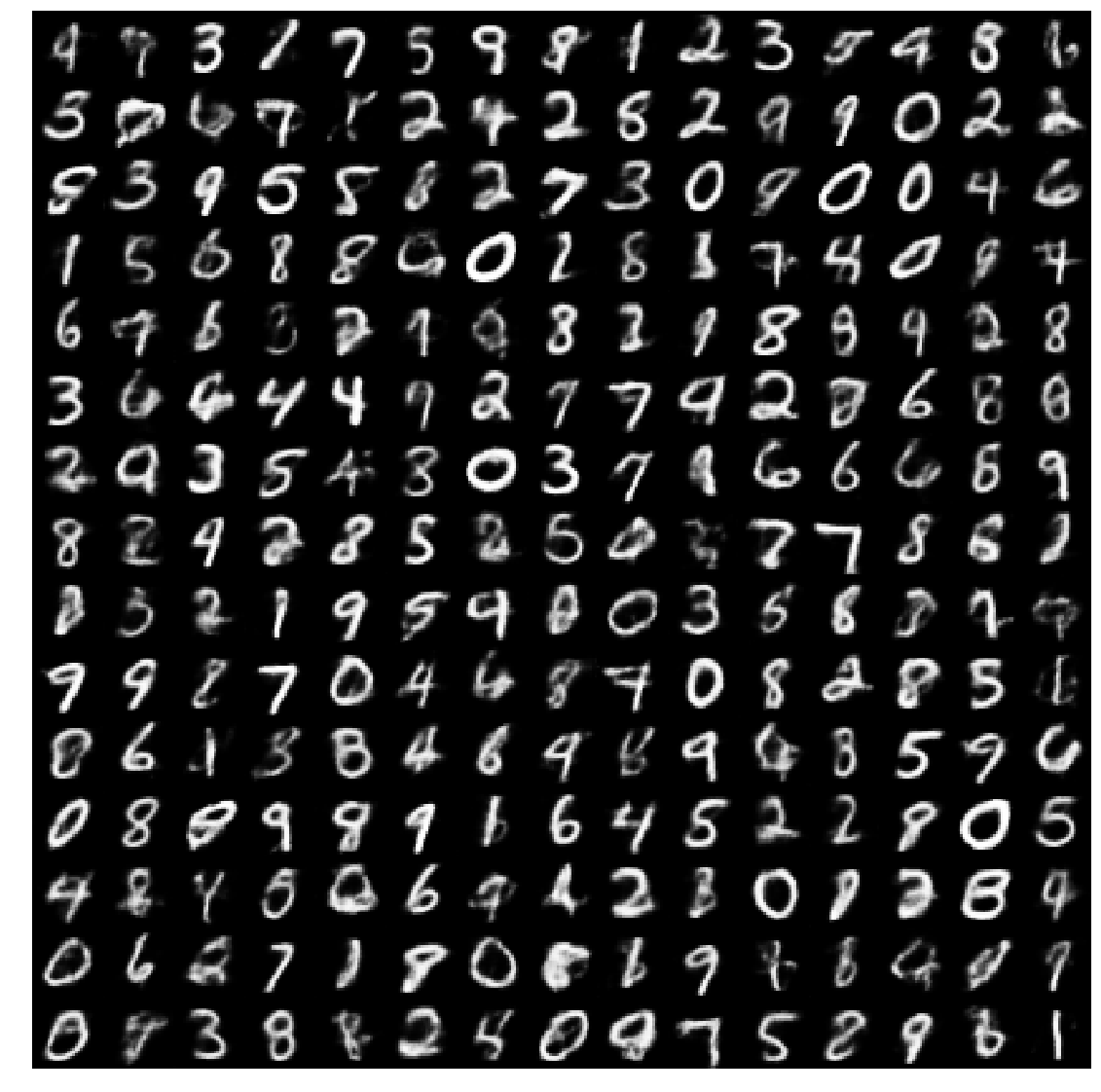} & \includegraphics[width=0.31\textwidth]{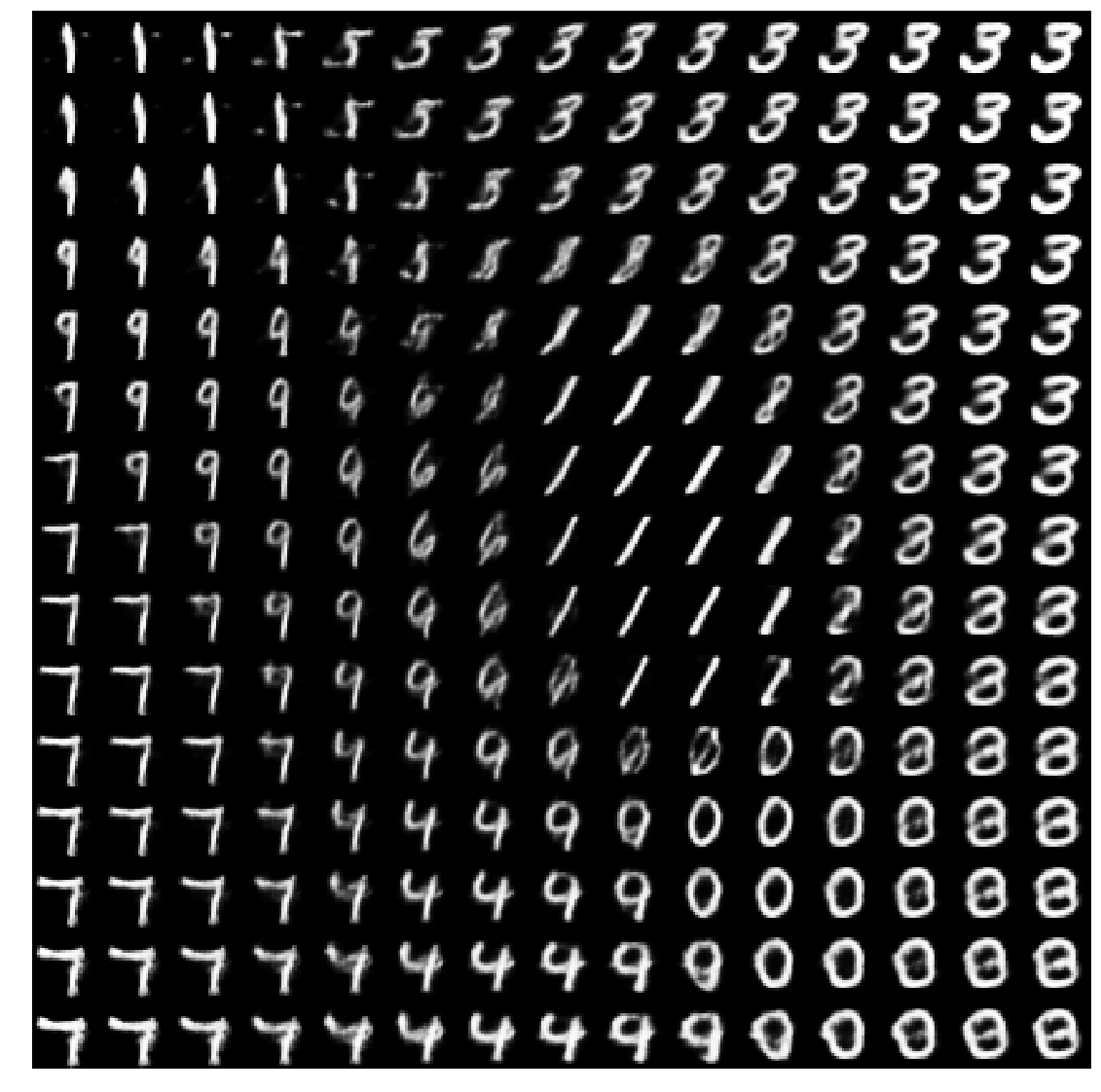}\tabularnewline
 & a) Test reconstruction & b) Random samples & c) Slice through code space\tabularnewline
\end{tabular}
\par\end{centering}
\caption{\label{fig:WAE-results}Qualitative results for WAE trained on MNIST.
In (a) odd rows are the real images.}
\end{figure}

\paragraph*{WAE results}

Here we present the results of training WAEs on MNIST dataset. The
architecture for the neural net is borrowed from RStudio's ``Keras
Variational Auto-encoder with Deconvolutions'' example\footnote{https://keras.rstudio.com/articles/examples/variational\_autoencoder\_deconv.html}.
This network has about 3.5M trainable parameters, almost an order
of magnitude less than the the 22M parameter network used by Tolstikhin
et al. \cite{WAE}. We consider three versions:
\begin{itemize}
\item CodeNorm---code normalization is used, the kernel width is kept fixed. 
\item Adaptive---no code normalization is used, kernel width is chosen
adaptively. This has two versions:
\begin{itemize}
\item AdaptiveBN---since code normalization can have other benefits (e.g.
improved optimization \cite{BatchNormHow}), we add batch normalization
as the initial layer of the decoder;
\item AdaptivePlain---no batch normalization layer added at all.
\end{itemize}
\end{itemize}
The CodeNorm version was trained for 60 epochs, but to allow the Adaptive
versions to reach a favorable configuration in the code space we trained
them for an extra 20 epochs at the initial learning rate. The latent
dimension is set to $d=8$ and all versions use the closed-form SMMD
penalty for fairness; further details are provided in the Appendix
\ref{sec:Code}.

Figure \ref{fig:WAE-results} (a)-(b) shows the reconstruction of
test images and random samples generated from Gaussian noise fed to
the decoder. We also take a planar slice through the origin in the
code space and feed the codes at the regular grid along this plane
into the decoder. Figure \ref{fig:WAE-results} (c) depicts the resulting
digit images, giving a taste of the manifold structure captured by
the models. Qualitatively, both of the Adaptive results are lower
quality than CodeNorm despite the former being trained for more epochs.
Quantitative results are presented in Table \ref{tab:WAEQuantitative}.
CodeNorm achieves the best test reconstruction error. We speculate
that the reason for this is that the gradient components of the MMD
penalty pointing in the direction of ``ideal'' samples (see Section
\ref{subsec:Code-Normalization}) add oscillations that hinder reduction
in the reconstruction loss of the Adaptive models. 

Next, we follow the suggestion of \cite{Cramer_Wold_AE} to compute
Mardia's multivariate skewness and normalized kurtosis statistics
of the latent code distribution of test data; we used the formulas
provided in \cite{Cramer_Wold_AE} and obtained the values as shown
in the table. We see that for both measures, the CodeNorm version
is better. Skewness is a measure of symmetry, so its small magnitude
indicates that the code distribution is symmetrically distributed
around the origin. Since kurtosis is a measure of outlier presence
\cite{kurtosis_RIP}, its small value indicates that there are no
outliers present in the code distribution. We verified experimentally
(not presented here) that code normalization is responsible for keeping
kurtosis under control. Indeed, removing the code normalization layer
from a trained network, modifying the latent layer incoming weights
so that the codes have zero mean and unit variance, and continuing
to train afterwards leads to increased kurtosis as predicted in Section
\ref{subsec:Code-Normalization}. 

Finally, we analyze the results using the B-statistic discussed in
Section \ref{sec:WAE-Training}. We computed the B-statistic using
$m=50$ batches of size $n=100$ from the test partition of MNIST.
The corresponding three sigma interval is $[-3/\sqrt{50},3/\sqrt{50}]=[-0.424,0.424]$.
Both CodeNorm and AdaptiveBN look good in terms of this statistic,
CodeNorm falling inside the interval; on the other hand AdaptivePlain
is somewhat farther away, indicating that its code distribution more
noticeably deviates from the target distribution. 

\setlength{\tabcolsep}{9pt}

\begin{table}
\begin{centering}
\begin{tabular}{lcccc}
\toprule 
WAE Version & Test MSE & Normalized Kurtosis & Skewness & B-Statistic\tabularnewline
\midrule 
CodeNorm & 0.0156 & -0.90 & 0.56 & 0.355\tabularnewline
AdaptiveBN & 0.0244 & 6.85 & 2.80 & 0.449\tabularnewline
AdaptivePlain & 0.0242 & 3.81 & 2.35 & 0.519\tabularnewline
\bottomrule
\end{tabular}
\par\end{centering}
\caption{\label{tab:WAEQuantitative}Quantitative comparison of different WAE
versions.}
\end{table}

\section{Conclusion}

We have introduced closed-form formulas for MMD, pointed out a relationship
with the BHEP statistic, and provided suggestions for WAE training
and monitoring. Our experiments confirm that the analytic formulation
improves over the stochastic approximation of the MMD, and demonstrate
that code normalization provides significant benefits when training
WAEs. An interesting avenue for future work is to investigate using
the unbiased MMD estimator for hypothesis testing instead of the BHEP
statistic. New analytic results would be needed to compute the variance
of this statistic for the composite null hypothesis of multivariate
normality and perhaps the asymptotic estimates from the MMD literature
can be used to determine testing thresholds. 

\bibliographystyle{plain}
\bibliography{biblio}

\appendix 

\section{Random Encoders}

\subsection{\label{subsec:Random-Encoders}MMD Formula}

In this section we consider Gaussian random encoders, where instead
of one code per input data point, we obtain a distribution of codes
given as $z_{i}\sim N(\mu_{i},\Sigma_{i}),i=1,2,...,n$. Here $n$
is the batch size, $\Sigma_{i}$ is a diagonal covariance matrix,
$\Sigma_{i}=\mathrm{diag}(\sigma_{i1}^{2},\sigma_{i2}^{2},...,\sigma_{id}^{2})$.
Both mean vectors $\mu_{i}\in\mathbb{R}^{d}$ and variance vectors
$\sigma_{i}\in\mathbb{R_{+}}^{d}$ are computed by applying neural
nets to the input data. Our goals is once again to obtain an estimator
for $\mathrm{MMD^{2}}(\mathcal{N}_{d}(\vec{0},I),Q)$. 

Note that the implied distribution of $Q$ for the current batch is
an equally weighted mixture of Gaussians $Q_{\mathrm{batch}}$ with
the distribution given by:
\[
q_{\mathrm{batch}}(z)\sim\frac{1}{n}\sum_{i=1}^{n}\prod_{k=1}^{d}\frac{e^{-(z_{k}-\mu_{ik})^{2}/(2\sigma_{ik}^{2})}}{\sqrt{2\pi\sigma_{ik}^{2}}},
\]
where we $z_{k}$ is the $k$-th component of the vector $z\in\mathbb{R}^{d}$.
We will replace sampling from $Q$ in the formula for MMD, Eq. (\ref{eq:mmd}),
by sampling from $Q_{\mathrm{batch}}$, and compute the second and
third terms in a closed form. Note that the first term depends only
on $P$ and will be the same as before; the computation of the remaining
terms is demonstrated in Section \ref{appendix:Random-Encoders},
and yields the following unbiased estimator:

\begin{align*}
\mathrm{MMD}_{u}^{2}(\mathcal{N}_{d}(\vec{0},I),Q_{\mathrm{batch}})= & \left(\frac{\gamma^{2}}{2+\gamma^{2}}\right)^{d/2}-\frac{2}{n}\sum_{i=1}^{n}\prod_{k=1}^{d}\left(\frac{\gamma^{2}}{1+\gamma^{2}+\sigma_{ik}^{2}}\right)^{1/2}e^{-\frac{\mu_{ik}{}^{2}}{2(1+\gamma^{2}+\sigma_{ik}^{2})}}+\\
+ & \frac{1}{n^{2}}\sum_{i=1}^{n}\sum_{j=1}^{n}\prod_{k=1}^{d}\left(\frac{\gamma^{2}}{\gamma^{2}+\sigma_{ik}^{2}+\sigma_{jk}^{2}}\right)^{1/2}e^{-\frac{(\mu_{ik}-\mu_{jk}){}^{2}}{2(\gamma^{2}+\sigma_{ik}^{2}+\sigma_{jk}^{2})}}.
\end{align*}

When the noise is isotropic, namely $\Sigma_{i}=\mathrm{diag}(\sigma_{i}^{2},\sigma_{i}^{2},...,\sigma_{i}^{2})$
with $\sigma_{i}\in\mathbb{R_{+}}$(note that $\sigma_{i}$ was a
vector in the general case above, but here it is a single number),
we can rewrite this formula in a simpler form:
\begin{align*}
\mathrm{MMD}_{u}^{2}(\mathcal{N}_{d}(\vec{0},I),Q)= & \left(\frac{\gamma^{2}}{2+\gamma^{2}}\right)^{d/2}-\frac{2}{n}\sum_{i=1}^{n}\left(\frac{\gamma^{2}}{1+\gamma^{2}+\sigma_{i}^{2}}\right)^{d/2}e^{-\frac{\Vert\mu_{i}\Vert{}^{2}}{2(1+\gamma^{2}+\sigma_{i}^{2})}}+\\
+ & \frac{1}{n^{2}}\sum_{i=1}^{n}\sum_{j=1}^{n}\left(\frac{\gamma^{2}}{\gamma^{2}+\sigma_{i}^{2}+\sigma_{j}^{2}}\right)^{d/2}e^{-\frac{\Vert\mu_{i}-\mu_{j}\Vert{}^{2}}{2(\gamma^{2}+\sigma_{i}^{2}+\sigma_{j}^{2})}}.
\end{align*}
Note that setting the variances $\sigma_{i}^{2}=0$ gives rise to
the deterministic encoders where $z_{i}=\mu_{i}$, and the resulting
estimator is the same as $\mathrm{MMD}_{b}^{2}$ and not $\mathrm{MMD}_{u}^{2}$.
The difference is that the last term in the unbiased deterministic
estimator includes an average over distinct pairs $(i,j),i\neq j$,
whereas for the unbiased random estimator the average runs over all
pairs $(i,j)$. The latter is appropriate here because when $\sigma_{i}^{2}\neq0$,
in Eq. (\ref{eq:mmd}) one can sample $y,y'\sim Q_{\mathrm{batch}}$
independently from the same component of the Gaussian mixture. Doing
so in the deterministic case would have resulted in a biased estimate:
essentially instead of the U-statistic we would have gotten the upwards
biased V-statistic.

\subsection{\label{subsec:Code-Normalization-1}Code Normalization}

Random encoders require a separate treatment of the mean and variance
network outputs. Namely, using the above notation, code normalization
is given in coordinate-wise manner by 
\[
\mu_{\cdot k}\rightarrow(\mu_{\cdot k}-\mathrm{Mean}_{k}(Q_{\mathrm{batch}}))/\mathrm{SD}_{k}(Q_{\mathrm{batch}}),\:\mathrm{and\quad}\sigma_{\cdot k}\rightarrow\sigma_{\cdot k}/\mathrm{SD}_{k}(Q_{\mathrm{batch}}),
\]
where subscript $k$ is used to refer to the $k$-th coordinate, $k=1,2,...,d$,
and $\mathrm{SD}_{k}(Q_{\mathrm{batch}})=\sqrt{\mathrm{Var}_{k}(Q_{\mathrm{batch}})}$.
Since $Q_{\mathrm{batch}}$ is a mixture of Gaussians, closed form
expressions for mean and variance are available: 
\[
\mathrm{Mean}_{k}(Q_{\mathrm{batch}})=\frac{1}{n}\sum_{i=1}^{n}\mu_{ik},\:\mathrm{and}\qquad\mathrm{Var}_{k}(Q_{\mathrm{batch}})=\frac{1}{n}\sum_{i=1}^{n}(\mu_{ik}^{2}+\sigma_{ik}^{2})-\left(\frac{1}{n}\sum_{i=1}^{n}\mu_{ik}\right)^{2}.
\]

\section{\label{sec:Hypothesis-Testing} Hypothesis Tests for Multivariate
Normality}

In this section we discuss hypothesis testing using $\mathrm{SMMD}^{2}$
and provide thresholds that can be useful when monitoring progress
of WAE code distribution convergence on a single batch level. Note
that the discussion in the main text was limited to multi-batch testing
setup which had a simple null distribution due to the CLT. Our initial
discussion is set in a broader manner so as to encompass general testing
for multivariate normality. 

We quickly review the hypothesis testing setting following \cite{HZ}
with some notational changes. Let $X_{1},X_{2},...,X_{n}\in\mathbb{R}^{d}$
be i.i.d. random vectors from some underlying distribution. The problem
is to test the hypothesis that the underlying distribution is a non-degenerate
$d$-variate normal distribution: $X_{i}\sim\mathcal{N}_{d}(\vec{\mu},\Sigma)$,
for some mean vector $\vec{\mu}$ and non-degenerate covariance matrix
$\Sigma$. Note that the population mean vector and covariance matrix
are not known. 

The test of multivariate normality proceeds as follows. Let $\bar{X}=n^{-1}\sum_{i}X_{i}$
be the sample mean, and $S=(n-1)^{-1}\sum_{i}(X_{i}-\bar{X})(X_{i}-\bar{X})^{T}$
be the sample covariance matrix. Assuming non-degeneracy, define the
centered and whitened vectors 
\[
Z_{i}=S^{-1/2}(X_{i}-\bar{X}).
\]
Now, the task of testing multivariate normality of $\{X_{i}\}_{i=1}^{n}$
reduces to the simpler problem of testing whether the underlying distribution
of $\{Z_{i}\}_{i=1}^{n}$ is $\mathcal{N}_{d}(\vec{0},I)$. 

While Henze-Zirkler test \cite{HZ} carries out this last step by
using the BHEP statistic, it can also be achieved by using the $\mathrm{SMMD}^{2}$.
One computes the $\mathrm{SMMD}^{2}$ statistic for the sample $\{Z_{i}\}_{i=1}^{n}$
and checks whether it is above the test threshold, and if so, the
null hypothesis gets rejected. The most straightforward way to compute
the threshold is to run a Monte Carlo simulation: sample $\{Z_{i}\}_{i=1}^{n}$
from the null distribution $\mathcal{N}_{d}(\vec{0},I)$, and compute
the corresponding $\mathrm{SMMD}^{2}$ value; repeat this many times
to obtain the empirical sampling distribution of the statistic and
use the $100\cdot(1-\alpha)$-th percentile as the threshold for the
$\alpha$-level test. 

However, this approach is problematic due to the treatment of the
nuisance parameters $\vec{\mu}$ and $\Sigma$: the same sample is
used both for estimating mean and covariance, and then for testing
(this is somewhat like training and testing on the same data). The
most apparent consequence is that one introduces dependencies within
$\{Z_{i}\}_{i=1}^{n}$, namely, $n^{-1}\sum_{i}Z_{i}=\vec{0}$ and
$(n-1)^{-1}\sum_{i}Z_{i}Z_{i}^{T}=I_{d}$, rendering it no longer
an i.i.d sample. Thus, when using the Monte Carlo approach with $\{Z_{i}\}_{i=1}^{n}$
sampled directly from $\mathcal{N}_{d}(\vec{0},I)$ we would end up
with a wrong null distribution and, so, with the wrong thresholds.
Henze-Zirkler test \cite{HZ} uses appropriate corrections to account
for the nuisance parameters when computing the moments under the null.
These moments are then used to obtain a log-normal approximation to
the null distribution. A similar path can be potentially taken with
the $\mathrm{SMMD}^{2}$ statistic, but for simplicity we will explain
how to correct the issue with Monte Carlo sampling.

To fix the problem, the computation of the null distribution should
proceed from samples that satisfy the dependency relationships mentioned
above. Fortunately, constructing such samples is easy: we sample $\{Z_{i}^{\mathrm{orig}}\}_{i=1}^{n}$
from $\mathcal{N}_{d}(\vec{0},I)$, then apply centering by the mean
and whitening by the sample covariance matrix. The resulting sample
$\{Z_{i}\}_{i=1}^{n}$ satisfies the relationships $n^{-1}\sum_{i}Z_{i}=\vec{0}$
and $(n-1)^{-1}\sum_{i}Z_{i}Z_{i}^{T}=I_{d}$. This centered and whitened
sample is used to compute the $\mathrm{SMMD}^{2}$ values and to obtain
the thresholds. To prove the correctness of this procedure one has
to show that there is a measure preserving and test statistic preserving
one-to-one mapping between these samples originating from $\mathcal{N}_{d}(\vec{0},I)$
and samples if they were to originate from $\mathcal{N}_{d}(\vec{\mu},\Sigma)$
with the true $\vec{\mu}$ and $\Sigma$. Using the non-degeneracy
of $\Sigma$, with some linear algebra one can show that indeed there
is such a mapping given by an orthogonal linear transformation, see
Section \ref{subsec:Correctness-of-the}. The matrix of this transformation
depends on $\Sigma$ only, making it measure preserving. Since $\mathrm{SMMD}^{2}$
is rotation-invariant, the resulting sampling distributions coincide.

Before proceeding, we would like to mention a modification of the
above test where the goal is to test whether the sample comes from
a normal distribution with a diagonal covariance. This is a test that
both checks each dimension for normality and establishes the independence
between the dimensions. When conducting the test only the diagonal
of the sample covariance matrix is computed and used for transforming
$\{X_{i}\}_{i=1}^{n}$ to $\{Z_{i}\}_{i=1}^{n}$. The corresponding
Monte Carlo procedure takes $\{Z_{i}^{\mathrm{orig}}\}_{i=1}^{n}$
from $\mathcal{N}_{d}(\vec{0},I)$, and applies centering by the mean
and scaling each dimension by its standard deviation (just like code
normalization).

Table \ref{tab:thresholds} displays the thresholds corresponding
to the $0.05$ level test, for sample size of $n=100$ for varying
dimensions and kernel scales (we have included dimensions $16$ and
$32$ to give an idea about the overall trend; one expects the test
to lose power with an increasing dimensionality \cite{Ramdas_decreasing_power}).
The column ``Sample Type'' indicates what processing was applied
to the original sample $\{Z_{i}^{\mathrm{orig}}\}_{i=1}^{n}$ from
$\mathcal{N}_{d}(\vec{0},I)$, if any. The ``Original'' thresholds
can be used for testing the following simple hypothesis: given a sample
$\{X_{i}\}_{i=1}^{n}$ we would like to test whether the underlying
distribution is $\mathcal{N}_{d}(\vec{0},I)$. The ``Centered+Scaled''
and ``Centered+Whitened'' rows give the correct thresholds for composite
nulls, i.e. testing whether $X_{i}\sim\mathcal{N}_{d}(\vec{\mu},\Sigma)$,
for unknown $\vec{\mu}$ and $\Sigma$. ``Centered+Scaled'' corresponds
to the case where $\Sigma$ is assumed to be diagonal, and ``Centered+Whitened''
correspond to the case of a general non-degenerate $\Sigma$. As expected,
dependencies within the sample shift the null distribution of $\mathrm{SMMD}^{2}$
to the left considerably; also see Figure 2 in the main text for side
by side histograms of ``Centered+Scaled'' versus ``Original''
null distributions. Therefore, using the original thresholds for composite
hypotheses would have resulted in tests that are rather liberal. 

\begin{table}
\begin{centering}
\begin{tabular}{llrrrrrr}
\toprule 
Dimension & Sample Type & $s=1$ & 1/2 & 1/4 & 1/8 & 1/16 & HZ\tabularnewline
\midrule
\midrule 
$d=1$ & Original & 1.97 & 1.97 & 1.95 & 1.92 & 1.91 & 1.98\tabularnewline
 & Centered+Scaled/Whitened & -0.13 & 0.32 & 0.75 & 1.05 & 1.24 & 0.22\tabularnewline
\midrule 
$d=2$ & Original & 1.93 & 1.94 & 1.90 & 1.85 & 1.83 & 1.90\tabularnewline
 & Centered+Scaled & -0.57 & -0.12 & 0.39 & 0.86 & 1.16 & 0.23\tabularnewline
 & Centered+Whitened & -0.79 & -0.33 & 0.22 & 0.72 & 1.06 & 0.03\tabularnewline
\midrule 
$d=4$ & Original & 1.90 & 1.87 & 1.83 & 1.79 & 1.76 & 1.83\tabularnewline
 & Centered+Scaled & -1.09 & -0.66 & -0.02 & 0.63 & 1.12 & 0.30\tabularnewline
 & Centered+Whitened & -1.60 & -1.25 & -0.56 & 0.24 & 0.85 & -0.16\tabularnewline
\midrule 
$d=8$ & Original & 1.85 & 1.83 & 1.80 & 1.77 & 1.74 & 1.75\tabularnewline
 & Centered+Scaled & -1.76 & -1.36 & -0.59 & 0.34 & 1.11 & 0.55\tabularnewline
 & Centered+Whitened & -2.63 & -2.57 & -1.88 & -0.60 & 0.58 & -0.30\tabularnewline
\midrule 
$d=16$ & Original & 1.81 & 1.80 & 1.77 & 1.74 & 1.78 & 1.78\tabularnewline
 & Centered+Scaled & -2.65 & -2.30 & -1.47 & -0.15 & 1.08 & 1.05\tabularnewline
 & Centered+Whitened & -4.00 & -4.47 & -4.17 & -2.31 & 0.00 & -0.03\tabularnewline
\midrule 
$d=32$ & Original & 1.77 & 1.77 & 1.74 & 1.71 & 1.76 & 1.30\tabularnewline
 & Centered+Scaled & -3.87 & -3.59 & -2.78 & -1.13 & 0.75 & 1.02\tabularnewline
 & Centered+Whitened & -5.86 & -7.12 & -7.91 & -5.72 & -1.24 & 0.06\tabularnewline
\bottomrule
\end{tabular}
\par\end{centering}
\caption{\label{tab:thresholds}Empirical thresholds for hypothesis tests with
size $\alpha=0.05$. See text for the details of when each kind of
threshold should be used. Here, $n=100$, kernel scale $s=\gamma^{2}/d$.
HZ is the $\gamma$ suggested by Henze and Zirkler \cite{HZ}, see
the main article for the formula.}
\end{table}

\paragraph*{Monitoring WAE Training Progress}

We can consider two ways of monitoring progress: at a single batch
and multi-batch levels; the multi-batch version is explained in Section
3.1 of the main text under the heading ``Monitoring WAE Training
Progress''. When inspecting the value of $\mathrm{SMMD}^{2}$ for
a single batch, one can use the above thresholds for hypothesis testing
as a guideline. Assuming that this batch is from validation or test
set, we can use the above thresholds listed in the ``Original''
rows of Table \ref{tab:thresholds}. By looking at these values, we
suggest using $2.0$ as an easy to remember liberal threshold. This
applies to code normalized batches as long as the normalization is
done using population statistics. However, when the batch is normalized
using its own statistics, then the appropriate thresholds are given
by the ``Centered+Scaled'' rows. We should stress again that even
upon convergence to the target distribution, one should still expect
oscillations of the $\mathrm{SMMD}^{2}$ values: it is not the case
that samples from the target distribution all have $\mathrm{SMMD}^{2}$
equal to zero, instead they follow the appropriate null distribution.

Of course, the single-batch approach above that treats the whole validation/test
set as one batch would result in a more powerful test. However, the
multi-batch B-statistic or E-statistic tests are simple to state and
are computationally inexpensive as they avoid constructing the large
pair-wise distance matrices for the overall test. Moreover, neural
network packages such as Keras provide these types of averages automatically
if one adds the corresponding quantity as a validation metric. At
a theoretical level, one should keep in mind that given enough power
we will always reject the null: with real-life data one rarely expects
the neural net to exactly reproduce the normal distribution. Rejecting
the null at high power does not mean that the distributions are easily
distinguishable: the practical difference can be so small that a classifier
trained to distinguish the two distributions (think of an adversary
from an adversarial WAE) would perform at a nearly chance level. Based
on these considerations, using the B-Statistic with $m=30-50$ should
be a reasonable choice, see the discussion in \cite[Appendix 1]{FRISTON20121300}
albeit in a different context; for power calculations for the MMD
based tests one can refer to \cite{Sutherland_model_criticismMMD}. 

\section{\label{sec:Derivations-and-Proofs}Derivations and Proofs}

\subsection{\label{appendix:Deterministic-Encoders}Deterministic Encoders}

We start with the expression 
\begin{equation}
\mathrm{MMD}_{u}^{2}(\mathcal{N}_{d},Q_{n})=\E_{x,x'\sim\mathcal{N}_{d}}[k(x,x')]-\frac{2}{n}\sum_{i=1}^{n}\E_{x\sim\mathcal{N}_{d}}[k(x,z_{i})]+\frac{1}{n(n-1)}\sum_{i=1}^{n}\sum_{j\neq i}^{n}k(z_{i},z_{j}).\label{eq:mmdu-1}
\end{equation}

and show that the first two expectations can be computed in closed
form. Let us start with the second term, and rewrite each summand
as an integral:
\begin{align}
\E_{x\sim P}[k(x,z)]= & \int_{\mathbb{R}^{d}}e^{-\Vert x-z\Vert^{2}/(2\gamma^{2})}(2\pi)^{-d/2}e^{-\Vert x\Vert^{2}/2}dx\nonumber \\
= & (2\pi\gamma^{2})^{d/2}\int_{\mathbb{R}^{d}}(2\pi\gamma^{2})^{-d/2}e^{-\Vert x-z\Vert^{2}/(2\gamma^{2})}(2\pi)^{-d/2}e^{-\Vert x\Vert^{2}/2}dx.\label{eq:temp2}
\end{align}
Since $\Vert x-z\Vert^{2}=\Vert z-x\Vert^{2}$, the integral in this
expression can be recognized as the probability density function of
the sum $Z=U+V$ where $U\sim N(\vec{0},\gamma^{2}I)$ and $V\sim N(\vec{0},I)$.
Being a sum of two normal distributions, adding means and variances
we get, $Z\sim N(\vec{0},(1+\gamma^{2})I)$, and the above expression
computes to 
\begin{equation}
\E_{x\sim P}[k(x,z)]=(2\pi\gamma^{2})^{d/2}(2\pi(1+\gamma^{2}))^{-d/2}e^{-\frac{\Vert z\Vert^{2}}{2(1+\gamma^{2})}}=\left(\frac{\gamma^{2}}{1+\gamma^{2}}\right)^{d/2}e^{-\frac{\Vert z\Vert^{2}}{2(1+\gamma^{2})}}.\label{eq:mmdu_xz}
\end{equation}

Next, we compute the first term in Eq. (\ref{eq:mmdu-1}) by rewriting
it as an integral:
\begin{align}
 & \E_{x,x'\sim P}[k(x,x')]=\nonumber \\
= & \int_{\mathbb{R}^{d}}\int_{\mathbb{R}^{d}}e^{-\Vert x-x'\Vert^{2}/(2\gamma^{2})}(2\pi)^{-d/2}e^{-\Vert x\Vert^{2}/2}(2\pi)^{-d/2}e^{-\Vert x'\Vert^{2}/2}dxdx'.\label{eq:mmdu_xx}
\end{align}
In this expression, let us replace $e^{-\Vert x'\Vert^{2}/2}$ by
$e^{-\Vert x'-w\Vert^{2}/2}$, and remember that we would get the
sought value by setting $w=\vec{0}$. Rewriting this as 
\begin{equation}
(2\pi\gamma^{2})^{d/2}\int_{\mathbb{R}^{d}}\left(\int_{\mathbb{R}^{d}}(2\pi\gamma^{2})^{-d/2}e^{-\Vert x-x'\Vert^{2}/(2\gamma^{2})}(2\pi)^{-d/2}e^{-\Vert x\Vert^{2}/2}dx\right)(2\pi)^{-d/2}e^{-\Vert x'-w\Vert^{2}/2}dx',\label{eq:mmdu_trick}
\end{equation}
With this replacement, we can recognize the inner integral as the
density function of the sum of two multivariate normal variables.
Interpreting the outer integral similarly, we can see that the entire
double integral captures the probability density function of the sum
$W=A+B+C$, where $A\sim N(\vec{0},\gamma^{2}I)$, $B\sim N(\vec{0},I)$
and $C\sim N(\vec{0},I)$. Being a sum of three normal distributions,
adding means and variances we get $W\sim N(\vec{0},(2+\gamma^{2})I)$,
immediately giving the expression for this integral as
\[
(2\pi(2+\gamma^{2}))^{-d/2}e^{-\frac{\Vert w\Vert^{2}}{2(2+\gamma^{2})}}
\]
Including the multiplier in front of the integral, and setting $w=\vec{0}$,
we obtain:
\[
\E_{x,x'\sim P}[k(x,x')]=\left(\frac{\gamma^{2}}{2+\gamma^{2}}\right)^{d/2}.
\]

Putting everything together we obtain the closed-form formula for
$\mathrm{MMD}_{u}^{2}(\mathcal{N}_{d},Q_{n})$.

\paragraph*{Variance}

Since its computation involves taking a random sample from $Q$, we
see that $\mathrm{MMD}_{u}^{2}$ is a random variable. Thus, even
when $Q=P=\mathcal{N}_{d}$, the estimator $\mathrm{MMD}_{u}^{2}(\mathcal{N}_{d},Q_{n})$
will not be identically zero. It is important to understand the behavior
of this random variable; using the hypothesis testing terminology,
we refer to this as the distribution of $\mathrm{MMD}_{u}^{2}$ under
the null---the null hypothesis being $Q=P=\mathcal{N}_{d}$. By unbiasedness,
we have that the null mean is zero: 
\[
\mathbb{E}[\mathrm{MMD}_{u}^{2}(\mathcal{N}_{d},Q_{n}))]=\mathrm{MMD}^{2}(\mathcal{N}_{d},\mathcal{N}_{d})=0,
\]
where the expectation is over various realizations of the sample $Q_{n}=\{z_{i}\}_{i=1}^{n}$
from $Q=\mathcal{N}_{d}$. This immediately means that in contrast
to $\mathrm{MMD}^{2}$, the estimator $\mathrm{MMD}_{u}^{2}$ can
take negative values.

Next, we would like to obtain the variance $\mathrm{MMD}_{u}^{2}$
under the null. First, we rewrite $\mathrm{MMD}_{u}^{2}$ by defining,
\[
h(z,z')=\left(\frac{\gamma^{2}}{2+\gamma^{2}}\right)^{d/2}-\left(\frac{\gamma^{2}}{1+\gamma^{2}}\right)^{d/2}e^{-\frac{\Vert z\Vert^{2}}{2(1+\gamma^{2})}}-\left(\frac{\gamma^{2}}{1+\gamma^{2}}\right)^{d/2}e^{-\frac{\Vert z'\Vert^{2}}{2(1+\gamma^{2})}}+e^{-\frac{\Vert z-z'\Vert^{2}}{2\gamma^{2}}},
\]
and noting that
\[
\mathrm{MMD}_{u}^{2}(\mathcal{N}_{d},Q_{n})=\frac{1}{n(n-1)}\sum_{i=1}^{n}\sum_{j\neq i}^{n}h(z_{i},z_{j}).
\]
Now according to \cite[Appendix B.3 ]{mmd} we have 
\[
\mathbb{E}\left[\left(\mathrm{MMD}_{u}^{2}(\mathcal{N}_{d},Q_{n})\right)^{2}\right]=\frac{2}{n(n-1)}E_{z,z'\sim\mathcal{N}_{d}}[h^{2}(z,z')].
\]
This expression can be computed in a closed form using manipulations
similar to those used for computing $\mathrm{MMD}_{u}^{2}$. Since
the mean of $\mathrm{MMD}_{u}^{2}$ under the null is $0$, the null
variance is equal to the second moment, and we obtain the formula,
\begin{align}
\mathrm{Var}(\gamma,d,n)\triangleq & \mathbb{E}\left[\left(\mathrm{MMD}_{u}^{2}(\mathcal{N}_{d},Q_{n})\right)^{2}\right]=\nonumber \\
= & \frac{2}{n(n-1)}\left[\left(\frac{\gamma^{2}}{2+\gamma^{2}}\right)^{d}+\left(\frac{\gamma^{2}}{4+\gamma^{2}}\right)^{d/2}-2\left(\frac{\gamma^{4}}{(1+\gamma^{2})(3+\gamma^{2})}\right)^{d/2}\right]\label{eq:Variance-1}
\end{align}

\subsection{\label{appendix:Random-Encoders}Random Encoders}

Let us start by computing the second term in Eq. (\ref{eq:mmd}),
namely

\begin{align}
\E_{x\sim P,y\sim Q_{\mathrm{batch}}}[k(x,y)] & =\int_{\mathbb{R}^{d}}\int_{\mathbb{R}^{d}}\left(\frac{1}{n}\sum_{i=1}^{n}\prod_{k=1}^{d}\frac{e^{-(y_{k}-\mu_{ik})^{2}/(2\sigma_{ik}^{2})}}{\sqrt{2\pi\sigma_{ik}^{2}}}\right)\cdot e^{-\frac{\Vert x-y\Vert^{2}}{2\gamma^{2}}}(2\pi)^{-d/2}e^{-\frac{\Vert x\Vert^{2}}{2}}dxdy\nonumber \\
= & \int_{\mathbb{R}^{d}}\int_{\mathbb{R}^{d}}\left(\frac{1}{n}\sum_{i=1}^{n}\prod_{k=1}^{d}\frac{e^{-(y_{k}-\mu_{ik})^{2}/(2\sigma_{ik}^{2})}}{\sqrt{2\pi\sigma_{ik}^{2}}}\right)\cdot\prod_{k=1}^{d}\frac{e^{-\frac{(x_{k}-y_{k}){}^{2}}{2\gamma^{2}}}e^{-\frac{x_{k}^{2}}{2}}}{\sqrt{2\pi}}dxdy\nonumber \\
= & \frac{1}{n}\sum_{i=1}^{n}\int_{\mathbb{R}^{d}}\int_{\mathbb{R}^{d}}\prod_{k=1}^{d}\frac{e^{-(y_{k}-\mu_{ik})^{2}/(2\sigma_{ik}^{2})}e^{-\frac{(x_{k}-y_{k}){}^{2}}{2\gamma^{2}}}e^{-\frac{x_{k}^{2}}{2}}}{\sqrt{2\pi\sigma_{ik}^{2}}\sqrt{2\pi}}dxdy,\label{eq:temp3}
\end{align}
where $y_{k}$ is the $k$-th coordinate of $y\in\mathbb{R}^{d}$.
Note that integrations over the dimensions of $\mathbb{R}^{d}$ are
independent, so the main component that we need to compute is 
\[
\sqrt{2\pi\gamma^{2}}\int_{\mathbb{R}}\int_{\mathbb{R}}\frac{e^{-(y_{k}-\mu_{ik})^{2}/(2\sigma_{ik}^{2})}e^{-\frac{(x_{k}-y_{k}){}^{2}}{2\gamma^{2}}}e^{-\frac{x_{k}^{2}}{2}}}{\sqrt{2\pi\sigma_{ik}^{2}\sqrt{2\pi\gamma^{2}}}\sqrt{2\pi}}dx_{k}dy_{k},
\]
where we multiplied and divided by the normalizing factor for the
kernel. Let us replace $\mu_{ik}$ in the first exponential by $w$,
and reasoning as with Eq. (\ref{eq:mmdu_trick}), we see that the
integral gives the probability density function of $W=A+B+C$, where
$A\sim N(0,\sigma_{ik}^{2})$, $B\sim N(0,\gamma^{2})$, and $C\sim N(0,1)$.
Thus, the integral is given by the pdf of $W\sim N(0,1+\gamma^{2}+\sigma_{ik}^{2})$.
Including the multiplier in front of the integral, and replacing $w=\mu_{ik}$,
we get:
\[
\left(\frac{\gamma^{2}}{1+\gamma^{2}+\sigma_{ik}^{2}}\right)^{1/2}e^{-\frac{\mu_{ik}{}^{2}}{2(1+\gamma^{2}+\sigma_{ik}^{2})}}.
\]
Putting this back into the last expression in Eq. (\ref{eq:temp3}),
we obtain
\[
\E_{x\sim P,y\sim Q_{\mathrm{batch}}}[k(x,y)]=\frac{1}{n}\sum_{i=1}^{n}\prod_{k=1}^{d}\left(\frac{\gamma^{2}}{1+\gamma^{2}+\sigma_{ik}^{2}}\right)^{1/2}e^{-\frac{\mu_{ik}{}^{2}}{2(1+\gamma^{2}+\sigma_{ik}^{2})}}.
\]

Next we will compute the third term in Eq. (\ref{eq:mmd}), namely
\begin{align*}
 & \E_{y,y'\sim Q_{\mathrm{batch}}}[k(y,y')]=\\
= & \int_{\mathbb{R}^{d}}\int_{\mathbb{R}^{d}}\left(\frac{1}{n}\sum_{i=1}^{n}\prod_{k=1}^{d}\frac{e^{-(y_{k}-\mu_{ik})^{2}/(2\sigma_{ik}^{2})}}{\sqrt{2\pi\sigma_{ik}^{2}}}\right)\cdot e^{-\frac{\Vert y-y'\Vert^{2}}{2\gamma^{2}}}\cdot\left(\frac{1}{n}\sum_{j=1}^{n}\prod_{k=1}^{d}\frac{e^{-(y'_{k}-\mu_{jk})^{2}/(2\sigma_{jk}^{2})}}{\sqrt{2\pi\sigma_{jk}^{2}}}\right)dydy'.
\end{align*}
As before, we can turn the exponential in the middle into a product
over the dimensions, and after distributing over the summations and
pushing the integrals into products, we obtain,
\[
\frac{1}{n^{2}}\sum_{i=1}^{n}\sum_{j=1}^{n}\prod_{k=1}^{d}\sqrt{2\pi\gamma^{2}}\int_{\mathbb{R}}\int_{\mathbb{R}}\frac{e^{-(y_{k}-\mu_{ik})^{2}/(2\sigma_{ik}^{2})}}{\sqrt{2\pi\sigma_{ik}^{2}}}\cdot\frac{e^{-\frac{(y_{k}-y'_{k})^{2}}{2\gamma^{2}}}}{\sqrt{2\pi\gamma^{2}}}\cdot\frac{e^{-(y'_{k}-\mu_{jk})^{2}/(2\sigma_{jk}^{2})}}{\sqrt{2\pi\sigma_{jk}^{2}}}dy_{k}dy_{k}'.
\]
In the double integral, let us replace $\mu_{ik}$ with $w$, keeping
$\mu_{jk}$ intact. Now the integral can be split to inner and outer
piece, and computed similarly to Eq. (\ref{eq:mmdu_trick}) as the
probability distribution of the sum of three one-dimensional Gaussians:
$W=A+B+C$, where $A\sim N(0,\sigma_{ik}^{2})$, $B\sim N(0,\gamma^{2}),$and
$C\sim N(\mu_{jk},\sigma_{jk}^{2}).$ We immediately get $W\sim N(\mu_{jk},\gamma^{2}+\sigma_{ik}^{2}+\sigma_{jk}^{2})$,
and the expression for the integral (multiplied by $\sqrt{2\pi\gamma^{2}}$)
in terms of $w$ is 
\[
\left(\frac{\gamma^{2}}{\gamma^{2}+\sigma_{ik}^{2}+\sigma_{jk}^{2}}\right)^{1/2}e^{-\frac{(w-\mu_{jk}){}^{2}}{2(\gamma^{2}+\sigma_{ik}^{2}+\sigma_{jk}^{2})}}.
\]
Substituting back $w=\mu_{ik}$ we obtain:
\[
\E_{y,y'\sim Q_{\mathrm{batch}}}[k(y,y')]=\frac{1}{n^{2}}\sum_{i=1}^{n}\sum_{j=1}^{n}\prod_{k=1}^{d}\left(\frac{\gamma^{2}}{\gamma^{2}+\sigma_{ik}^{2}+\sigma_{jk}^{2}}\right)^{1/2}e^{-\frac{(\mu_{ik}-\mu_{jk}){}^{2}}{2(\gamma^{2}+\sigma_{ik}^{2}+\sigma_{jk}^{2})}}.
\]
Collecting all the terms in Eq. (\ref{eq:mmd}), we obtain the formula
for $\mathrm{MMD}_{u}^{2}(\mathcal{N}_{d},Q_{\mathrm{batch}})$ in
Section \ref{subsec:Random-Encoders}.

\subsection{\label{subsec:Correctness-of-the}Correctness of the Monte Carlo
Sampling Procedure}

Here we prove the existence of an orthogonal matrix that establishes
a one-to-one measure-preserving mapping between centered-whitened
samples from $\mathcal{N}_{d}(\vec{\mu},\Sigma)$ and $\mathcal{N}_{d}$.
Consider the diagonalization of the true covariance matrix $\Sigma=ODO^{T}$,
where $D$ is a diagonal, and $O$ is an orthogonal matrix---this
is possible by the symmetry of $\Sigma$. Given a sample $X_{i}\sim\mathcal{N}_{d}(\vec{\mu},\Sigma)$
we can write $X_{i}=\vec{\mu}+OD^{1/2}Y_{i}$, where $Y_{i}$ distributed
as $\mathcal{N}_{d}$. 

Let $Z_{i}^{X}=S_{X}^{-1/2}(X_{i}-\bar{X})$ be the centered-whitened
$X_{i}$, and let $Z_{i}^{Y}=S_{Y}^{-1/2}(Y_{i}-\bar{Y})$ be the
centered-whitened $Y_{i}$. We will show that $Z_{i}^{X}=RZ_{i}^{Y}$
for some orthogonal matrix (i.e. rotation) $R$ computed below.

We start by noting that $\bar{X}=\vec{\mu}+OD^{1/2}\bar{Y}$, and
that the following relationship holds between the sample variance
matrices:
\begin{align}
S_{X} & =(n-1)^{-1}\sum_{i}(X_{i}-\bar{X})(X_{i}-\bar{X})^{T}\nonumber \\
= & (n-1)^{-1}\sum_{i}(\vec{\mu}+OD^{1/2}Y_{i}-\vec{\mu}-OD^{1/2}\bar{Y})(\vec{\mu}+OD^{1/2}Y_{i}-\vec{\mu}-OD^{1/2}\bar{Y})^{T}\label{eq:sxsy}\\
= & (n-1)^{-1}\sum_{i}OD^{1/2}(Y_{i}-\bar{Y})(Y_{i}-\bar{Y})^{T}(OD^{1/2})^{T}\nonumber \\
= & OD^{1/2}\left[(n-1)^{-1}\sum_{i}(Y_{i}-\bar{Y})(Y_{i}-\bar{Y})^{T}\right](OD^{1/2})^{T}=OD^{1/2}S_{Y}D^{1/2}O^{T}.\nonumber 
\end{align}
Now we have, 

\begin{align*}
Z_{i}^{X} & =S_{X}^{-1/2}(X_{i}-\bar{X})=S_{X}^{-1/2}(\vec{\mu}+OD^{1/2}Y_{i}-\vec{\mu}-OD^{1/2}\bar{Y})\\
= & S_{X}^{-1/2}OD^{1/2}(Y_{i}-\bar{Y})=S_{X}^{-1/2}OD^{1/2}S_{Y}^{1/2}S_{Y}^{-1/2}(Y_{i}-\bar{Y})\\
= & RS_{Y}^{-1/2}(Y_{i}-\bar{Y})=RZ_{i}^{Y}.
\end{align*}
To finish the proof, we need to show that $R=S_{X}^{-1/2}OD^{1/2}S_{Y}^{1/2}$
is an orthogonal matrix. It is enough to show that $RR^{T}=I$, and
indeed:
\begin{align*}
RR^{T} & =S_{X}^{-1/2}OD^{1/2}S_{Y}^{1/2}S_{Y}^{1/2}D^{1/2}O^{T}S_{X}^{-1/2}=S_{X}^{-1/2}(OD^{1/2}S_{Y}D^{1/2}O^{T})S_{X}^{-1/2}\\
= & S_{X}^{-1/2}S_{X}S_{X}^{-1/2}=I,
\end{align*}
where we used that both of the sample variance matrices are symmetric,
and replaced the expression in the parenthesis by $S_{X}$ based on
Eq. (\ref{eq:sxsy}).

\section{\label{sec:Code}Code}

\subsection{\label{appendix:mmd-python}Python Implementation of $\mathrm{SMMD}^{2}$ }

This implementation uses Tensorflow \cite{tensorflow2015-whitepaper}.
The choice of the kernel width in the adaptive case can be explained
as follows. As the neural network converges to the normal distribution,
the quantity \texttt{mean\_norms2}, which captures the mean squared
distance to the origin, converges to the latent dimension. This follows
from the fact that the mean of $\chi_{d}^{2}$ distribution is $d$.
As a result, in this limit, the adaptive and the fixed kernel width
would be approximately equal. The reason for using the distance to
the origin, and not to the center of mass, is to prioritize convergence
of the codes close to the origin. We use a non-robust statistic to
have a better control over the outliers.

\begin{lstlisting}[language=Python, caption={}]
#When CodeNorm is not used, set adaptive=True
def smmd(z, scale=1./8., adaptive=False):
    nf = tf.cast(tf.shape(z)[0], "float32")
    latent_dim = tf.cast(tf.shape(z)[1], "float32")
    
    norms2 = tf.reduce_sum(tf.square(z), axis=1, keepdims=True)
    dotprods = tf.matmul(z, z, transpose_b=True)
    dists2 = norms2 + tf.transpose(norms2) - 2. * dotprods
    
    if adaptive:
        mean_norms2 = tf.reduce_mean(norms2)
        gamma2 = tf.stop_gradient(scale*mean_norms2)
    else:
        gamma2 = scale*latent_dim
    
    variance = (gamma2/(2.+gamma2))**latent_dim + \
               (gamma2/(4.+gamma2))**(latent_dim/2.) - \
               2.*(gamma2**2./((1.+gamma2)*(3.+gamma2)))**(latent_dim/2.)
    variance = 2. * variance/(nf*(nf-1.))
    variance_normalization = (variance)**(-1./2.)
    
    Ekzz = (tf.reduce_sum(tf.exp(-dists2/(2.*gamma2))) - nf)/((nf * nf - nf))
    Ekzn = (gamma2/(1.+gamma2))**(latent_dim/2.)*\
           tf.reduce_mean(tf.exp(-norms2/(2.*(1.+gamma2))))
    Eknn = (gamma2/(2.+gamma2))**(latent_dim/2.)
    
    return variance_normalization*(Ekzz - 2.*Ekzn + Eknn)
\end{lstlisting}

\subsection{\label{subsec:Neural-net-architectures}WAE Architecture and Training}

For the MNIST WAE experiments we use the architecture below. Here
\texttt{CodeNorm} is the code normalization layer. It can be replaced
by \texttt{BatchNormalization(center=False, scale=False)} in Keras,
with the caveat that the computation of the batch variance in Keras
uses the biased sample estimate; to correct for this it needs to be
multiplied by $n/(n-1)$. 

The latent dimension is $d=8$, kernel scale is $s=1/8$, and the
regularization weight is $\lambda=0.01$. Adam \cite{ADAM} is used
for 60 epochs with default parameters except for the learning rate.
The learning rate for the first 20 epochs is 0.001, then set to 0.001/4
for the next 20, and set to 0.001/16 for the last 20 epochs. No regularization,
dropout, noise, or augmentation is used. Adaptive versions use \texttt{smmdu}
with \texttt{adaptive=True}, with further modifications as described
in the main text. These are trained for 80 epochs, with 20 extra epochs
at the initial rate.

\begin{lstlisting}[language=Python, caption={}]
#encoder
img_in = Input((28,28,1))
temp = Conv2D(filters = 1, kernel_size = (2, 2), strides = (1, 1),
              padding = "same", activation = "relu")(img_in) 
temp = Conv2D(filters = 64, kernel_size = (2, 2), strides = (2, 2),
              padding = "same", activation = "relu")(temp)
temp = Conv2D(filters = 64, kernel_size = (3, 3), strides = (1, 1),
              padding = "same", activation = "relu")(temp)
temp = Conv2D(filters = 64, kernel_size = (3, 3), strides = (1, 1),
              padding = "same", activation = "relu")(temp) 
temp = Flatten()(temp)
temp = Dense(units = 128, activation = "relu")(temp)
temp = Dense(units = latent_dim, activation = "linear")(temp)
code_out = CodeNorm()(temp)
encoder = Model(inputs=[img_in],outputs=[code_out])

# decoder
code_in = Input((latent_dim,))
temp = Dense(units = 128, activation = "relu")(code_in)
temp = Dense(units = 14*14*64, activation = "relu")(temp)
temp = Reshape(target_shape = (14, 14, 64))(temp)
temp = Conv2DTranspose(filters = 64, kernel_size = (3, 3), strides = (1, 1), 
              padding = "same", activation = "relu")(temp)
temp = Conv2DTranspose(filters = 64, kernel_size = (3, 3), strides = (1, 1), 
              padding = "same", activation = "relu")(temp)
temp = Conv2DTranspose(filters = 64, kernel_size = (3, 3), strides = (2, 2), 
              padding = "valid", activation = "relu")(temp)
img_out = Conv2D(filters = 1, kernel_size = (2, 2), strides = (1, 1), 
              padding = "valid", activation = "sigmoid")(temp)
decoder = Model(inputs=[code_in],outputs=[img_out])

# end-to-end WAE
x = Input((28,28,1))
z = encoder(x)
y = decoder(z)
wae = Model(inputs=[x],outputs=[y])

def wae_loss(x,y):
    return mean_squared_error(x,y) + 0.01*smmd(z, 1./8., adaptive=False)
    
wae.compile(optimizer='adam', loss=wae_loss)

\end{lstlisting}
\end{document}